%% file: main.tex
\documentclass[10pt,twocolumn,letterpaper]{article}

\usepackage{cvpr}              
\input{preamble}
\definecolor{cvprblue}{rgb}{0.21,0.49,0.74}
\usepackage[pagebackref,breaklinks,colorlinks,allcolors=cvprblue]{hyperref}

\def\paperID{2495} 
\def\confName{CVPR}
\def\confYear{2026}
\usepackage{amsthm}

\usepackage{booktabs}
\usepackage{times}
\usepackage{epsfig}
\usepackage{graphicx}
\usepackage{amsmath}
\usepackage{amssymb}
\usepackage{multirow}
\usepackage{tikz}
\usepackage{tikz-cd}
\usepackage{bm}
\usetikzlibrary{arrows,automata}
\usetikzlibrary{shadows}
\usetikzlibrary{plotmarks}
\usetikzlibrary{shapes}
\usepackage{array}
\usepackage{booktabs} 
\usepackage{colortbl}
\usepackage{xcolor}
\usepackage{soul}
\usepackage{arydshln}
\usepackage{mathtools}
\usepackage{fontawesome5}
\usepackage{makecell} 
\newcolumntype{?}{!{\vrule width 1pt}}
\newcolumntype{|}{!{\vrule width .5pt}}
\definecolor{darkred}{rgb}{0.6148, 0., 0.}
\definecolor{lightyellow}{rgb}{1., 1., 0.95}
\definecolor{lightgrape}{rgb}{0.93, 0.89, 0.98}
\definecolor{steelblue}{rgb}{0.2745,0.5098,0.7059}
\definecolor{lightcrimson}{rgb}{0.9176,0.4471,0.5412}

\usepackage[T1]{fontenc}

\DeclareRobustCommand{\piold}{\ensuremath{\pi_{\bm{\theta}_{\texttt{\textbf{old}}}}}\xspace}
\DeclareRobustCommand{\piold}{\ensuremath{\pi_{\bm{\theta}_{\texttt{\textbf{old}}}}}\xspace}
\newcommand{\roundval}[1]{^{\texttt{\textbf{(#1)}}}}

\title{
 \vspace{-5pt}
From Failure to Feedback: Group Revision Unlocks Hard Cases in \\ Object-Level Grounding
\vspace{-20pt}}

\author{
\parbox{\linewidth}{\centering Yuyuan Liu \textsuperscript{\rm 1}  $\quad$ Yiping Ji \textsuperscript{\rm 2} $\quad$ Anjie Le \textsuperscript{\rm 1} $\quad$ Jiayuan Zhu \textsuperscript{\rm 1} $\quad$ Jiazhen Pan \textsuperscript{\rm 3} $\quad$ \\ Can Peng \textsuperscript{\rm 1} $\quad$ Jiajun Deng \textsuperscript{\rm 4} $\quad$ Fengbei Liu \textsuperscript{\rm 5}$^{\text{(\faEnvelope[regular])}}$ $\quad$ Junde Wu \textsuperscript{\rm 1}
\\   \vspace{10pt}
\small  \textsuperscript{\rm 1} Department of Engineering Science, University of Oxford
 $\quad$   \textsuperscript{\rm 2} Australian Institute for Machine Learning, Adelaide University \\
 \textsuperscript{\rm 3} Technical University of Munich  $\quad$  
 \textsuperscript{\rm 4} University of Science and Technology of China  $\quad$  
 \textsuperscript{\rm 5} Cornell University }
 }
 
\begin{document}
\maketitle
\input{sec/0_abstract}    
\input{sec/1_intro}

\input{sec/2_literature}
\input{sec/3_method}

\input{sec/4_experiment}
\input{sec/5_conclusion}

{
    \small
    \bibliographystyle{ieeenat_fullname}
    \bibliography{main}
}


\end{document}

%% file: sec/0_abstract.tex
\begin{abstract}
Finetuning Large Vision-Language Models with reinforcement learning has emerged as a promising approach to enhance their capability in object-level grounding. However, existing methods, mainly based on GRPO, assign rewards at the response level. 
Such sparse reward, often criterion-induced, leads to minimal learning signals when all candidate responses fail in challenging scenarios.
In this work, we propose a group-revision optimisation paradigm that enhances learning on hard cases. 
It begins with a sampled initial response and generates a set of revised candidates to explore improved grounding outcomes. 
Inspired by reward shaping, we introduce a consolidation process that quantifies each candidate’s improvement over the initial attempt and converts it into informative shaping signals.
These signals are used to both refine the reward and modulate the advantage, amplifying the influence of high-quality revisions.
Our method achieves consistent gains across referring and reasoning segmentation, REC, and counting benchmarks compared with prior GRPO-based models. Our code is available at \url{https://github.com/yyliu01/GroupRevision}.
\vspace{-10pt}
\end{abstract}

%% file: sec/1_intro.tex
\section{Introduction}
\label{sec:intro}
\indent Large Vision–Language Models (LVLMs)~\cite{bai2025qwen2, wang2024qwen2, comanici2025gemini} have made significant advances in multimodal reasoning. 
From open-ended scene summarisation~\cite{liu2023visual, zhu2023minigpt, li2022blip} to content-based retrieval~\cite{chen2025janus, wu2025qwenimagetechnicalreport}, they have demonstrated strong performance in general image understanding.
Yet, when questions target specific objects within an image, such as their \textit{location}, LVLMs often struggle to ground answers precisely~\cite{leng2024mitigating, ji2023survey, rohrbach2018object}.
This reveals a broader limitation: current models fail to fully exploit the \textit{object-level} grounding signals present in the scene~\cite{hsieh2023sugarcrepe, liu2024survey, wang2025mixture}, rather than general visual reasoning. 
As a result, in tasks such as embodied navigation~\cite{windecker2025navitrace,qiao2025navbench,wang2025rethinking}, robotic manipulation~\cite{zhao2025manipbench,zhang2025chain,huang2025roboground}, and clinical imaging~\cite{kurz2025benchmarking}, where both  reasoning and object-level grounding are required, current LVLMs still fall short in delivering reliable performance.
\input{images/first_page_demo/demo}

\indent 
A promising direction to bridge this grounding gap is to fine-tune LVLMs using object-level reward signal via Group Relative Policy Optimisation (GRPO)~\cite{shao2024grpo}.
For a image-question pair, the policy samples a group of candidate responses, each scored based on whether it correctly localises the target object.
The policy is then updated to increase the likelihood of higher-reward outputs. 
In this vein, early efforts~\cite{liu2025visual, shen2025vlm, liu2025seg} trained LVLMs to predict object bounding boxes for detection~\cite{liu2025visual} or referring expression comprehension (REC)~\cite{shen2025vlm}, and to generate points (inside the object) that guide SAM2~\cite{ravi2024sam2} in producing segmentation masks~\cite{you2025segr1, huang2025samr1}. Recent work~\cite{liu2025visionreasoner} further shows that this training paradigm consistently improves performance on downstream tasks~\cite{lai2024lisa, paiss2023teaching} in multi-object scenarios.
\\
\indent Despite this progress, these GRPO-based methods 
still struggles to identify objects in complex scenes, especially when resolving references to similar-looking instances or interpreting fine-grained spatial relations. 
This mainly stems from their optimisation process: when challenging examples lead all responses in the group to fail the reward criterion, the policy receives minimal training signal, leaving these hard cases unlearned. 
This failure mode arises from GRPO’s sparse reward design~\cite{lightman2023lets, zhang2025lessons, mroueh2025revisiting}, where only the final output is evaluated, and no reward or guidance is given for the intermediate decision that leads to it.  Similar limitations have been observed in code generation~\cite{li2025codeprm, dai2024process} and math reasoning~\cite{zhang2025lessons}, motivating the development of Process Reward Models (PRMs)~\cite{zheng2025survey} that deliver feedback at each step toward the solution.
However, applying such step-wise reward to object-level visual grounding is challenging. It is non-trivial to link each intermediate reasoning sentence in the model’s Chain-of-Thought (CoT) to a concrete object in the image that can provide learning feedback.
This leads to a fundamental chicken-and-egg dilemma: the model needs more accurate intermediate guidance to obtain informative rewards, but these rewards are essential for learning to reason in the first place. 
\\
\indent Yet, a failed attempt doesn't have to be the end—but a clue. Although explicit step-wise feedback is unavailable, failed attempts often reveal what the model missed.
In such cases, revising the failure and exploring alternative responses can guide the model to attend to previously overlooked visual cues. 
As illustrated in Fig.~\ref{fig:first_page_demo}, a directly sampled \textbf{GRPO} group fails to correctly localise the referent. 
In this case, we wrap one failed response with a follow-up revision prompt and sample a group of revised responses based on it.
This \textbf{revision process} helps the model reinterpret the scene, allowing two candidates successfully localise the target.
Importantly, such revision transforms a previously unrewarded case into one that satisfies the standard success criterion (e.g., box\_IoU > 0.5~\cite{liu2025visual, liu2025seg, liu2025visionreasoner}), allowing it to contribute to training.
This case is not isolated. The \textcolor{darkred}{red curve} in Fig.~\ref{fig:first_page_demo} \textbf{(Quantitative study)} highlights hard training samples that 
receive no supervision signal from a group of responses and remain unlearned from GRPO.
After applying group revision, many of these samples recover reward.
The majority of \textcolor{steelblue!80}{improve above the IoU cutoff}, indicating consistent benefits across the dataset from the revised results. \\
\indent While the revision group often yields better responses, they are evaluated independently of the initial failure.  
As a result, the policy receives no credit assignment for what actually changed to fix the error.  
Without this connection, the model treats each outcome as isolated, missing the opportunity to learn what improvements truly mattered.  
To bridge this gap, a consolidation process is needed to turn these improvements into measurable learning signals. It enables the model not just to retry, but to learn \textit{why it succeeds after failing}.  
By internalising such improvements, the model can ultimately produce more accurate grounding results.
\\
\indent In this work, we introduce a group revision optimisation paradigm that enables LVLMs to recover training signal from challenging hard cases.  
Unlike prior methods that sample responses to answer the query, our approach forms a candidate group to collaboratively revise an earlier response.
To achieve this, we first sample a single response from the policy given a image–question pair.  
We then generate a revision group conditioned on this response, aiming to better localise ambiguous or hard-to-detect objects.  
To transfer the gains from group revision into learnable training signals, we introduce a consolidation process inspired by reward shaping~\cite{ng1999policy, christiano2017deep,wiewiora2003potential}.
In this process, we compute an alignment cost that quantifies how much the revision improves upon the initial response, relative to the ground-truth objects. 
This alignment cost is then converted into a shaping signal that assigns finer-grained rewards within the revision group, allowing the model to identify which candidates are truly helpful.
We further modulate the advantage with this shaping signal, so that informative candidates have a stronger impact on learning.
As a result, the model is better guided to internalise improvements from revision, leading to more effective learning from difficult cases. 
In summary, we make the following core contributions:
\begin{itemize}
\item We propose a group-revision training paradigm, which revises an initial attempt through a group of responses, to unlock learning signals from hard cases in LVLMs;
\item Within this strategy, we propose a consolidation process to quantify improvements over the initial output, resulting in a shaping signal used to construct the rewards; and
\item This signal is further used to scale the advantage, boosting the contribution of informative candidates in training.
\end{itemize}
We evaluate our method on multiple visual perception tasks and observe robust improvements over GRPO-based state-of-the-art (SoTA) methods. Notably, our approach achieves average gains of {+2.16\%}~\cite{lai2024lisa} on reasoning segmentation, {+2.22\%} on referring segmentation~\cite{yu2016modeling}, as well as {+4.27\%} on object counting~\cite{paiss2023teaching, deitke2025molmo} tasks.

%% file: images/first_page_demo/demo.tex
\begin{figure}[t!]
    \centering
    \includegraphics[width=\linewidth]{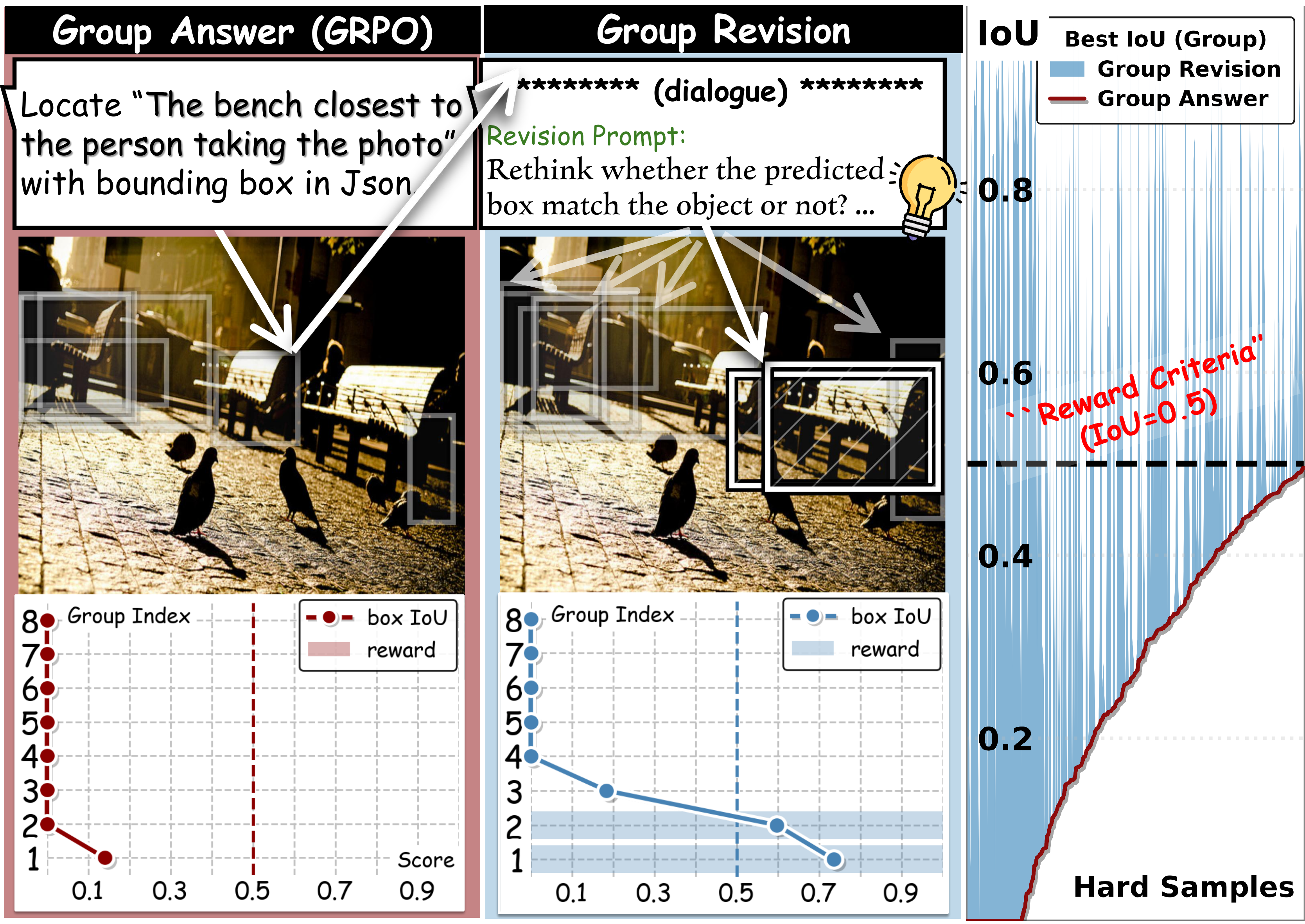}
    \caption{\textbf{Group Answer (GRPO)}: A group of responses (n=8) from Qwen2.5VL-7B~\cite{bai2025qwen2} yields unsatisfactory grounding, with the best box IoU only 14.72\%.
\textbf{Group Revision}: A group of responses that revises the mistaken answer improves the best IoU to 74.57\%.
\textbf{Quantitative study:} The \textcolor{darkred}{red curve} shows the best IoU within each GRPO group on the hard cases through the training set, while the \textcolor{steelblue}{group revision} effectively recover the reward signal.}
    \label{fig:first_page_demo}
    \vspace{-15pt}
\end{figure}

%% file: sec/2_literature.tex
\section{Related work}
\textbf{Large Vision-Language Models (LVLMs)} (also known as Multimodal Large Language Models) first gained attention with  Flamingo~\cite{alayrac2022flamingo} by DeepMind, which injects sampled visual tokens into the hidden layers of a frozen language model. For the first time, this enabled a language model to engage in image-grounded conversations. 
Motivated by this insight, subsequent works~\cite{he2024ma, zhu2023minigpt, driess2023palm} either align visual and language tokens using pretrained transformer-based adapters~\cite{li2023blip, zhu2023minigpt, he2024ma, driess2023palm}, or map CLIP-derived features into the language embedding space via visual instruction tuning~\cite{liu2023visual, wang2024qwen2, bai2025qwen2, chen2024expanding, zhu2025internvl3, wang2025visualprm}. 
Despite their strong general performance, LVLMs often struggle with instance-level queries~\cite{liu2024survey, rohrbach2018object, biten2022let}. They tend to hallucinate details~\cite{leng2024mitigating, ji2023survey}, and fail to generate accurate, object-aligned outputs (e.g., masks or boxes) for the referenced entities.
These issues remain a well-known obstacle to deploying LVLMs in tasks that demand precise visual perception~\cite{bai2024hallucination, wang2025rethinking, windecker2025navitrace,liu2025auralsam2,zhang2025chain,huang2025roboground}. \\
\textbf{Supervised Fine-Tuning (SFT)} was initially proposed to alleviate this issue. To move from image-level grounding to precise pixel-level perception, a common strategy~\cite{lai2024lisa, xia2024gsva, rasheed2024glamm, kirillov2023segment} is to let LVLMs emit special tokens that trigger pretrained segmentation models~\cite{ravi2024sam2} to produce query-aligned masks. Such `embedding-as-mask' paradigm has also been explored for detection~\cite{wu2024visionllm}, visual grounding~\cite{yu2025vpt}, video tracking~\cite{bai2024videolisa, yan2024visa}, and for depth estimation~\cite{bigverdi2025perceptiontokens}.
However, their benefits of strong supervision come with a trade-off, whereby SFT can overfit LVLMs to the supervised domain and erode foundation level generalisation~\cite{liu2025seg, zang2024overcoming, zhang2024vision}. This degradation is often accompanied with weakened chain-of-thought~\cite{chung2024scaling}, undermining the model’s reasoning process.
\\
Instead, \textbf{Reinforcement Fine-Tuning (RFT)} do not require the heavy supervision signal~\cite{zhou2025reinforcedMLLM}. Rather than memorising token-level answers, RFT optimises behaviour via task-specific rewards and constraining updates with KL regularisation~\cite{ouyang2022instructgpt, yu2025dapo}. As a result, the LVLM’s original capabilities are largely preserved.
Thanks to open-sourced foundation models~\cite{wang2024qwen2, bai2025qwen2, zhu2025internvl3, chen2025janus}
and the critic-free, GPU-friendly design of GRPO~\cite{shao2024grpo, guo2025deepseekr1}, RFT has gained growing interest, with a wide range of follow-up works for the post-training~\cite{pan2025medvlmr1, jiang2025rexthinker}. 
Recent efforts focus on visual grounded reasoning~\cite{cao2025groundr1,xu2025medgroundr1,zhou2025guig1,ye2025guiarp,bai2025univg,feng2025video,yang2025look}, which optimises textual responses and implicitly adjusts bounding boxes to localise the supporting evidence. 
However, these methods lack explicit object-level learning signal, making them difficult to guarantee fine-grained grounding during reasoning.
Our work follows a distinct line of research that directly optimises the policy to generate bounding boxes~\cite{liu2025visual}, predict points~\cite{huang2025samr1, you2025segr1}, or handle counting tasks under multi-object scenarios~\cite{liu2025visionreasoner}.  
While previous methods struggle with hard samples when all generated responses fail to identify the correct object, limiting the policy optimisation.
In contrast, our approach treats the initial response as an intermediate step and generates a group of follow-up candidates to collaboratively revise it.
This group-based revision strategy better handles ambiguous references and enables the model to recover informative signals even in challenging cases. \\
\textbf{Reward Shaping}~\cite{andrychowicz2017hindsight, christiano2017deep, ng1999policy} is commonly used to mitigate sparse or delayed supervision by providing denser learning feedback. Potential–based shaping~\cite{ng1999policy} formalises this idea and has been widely adopted as a principled way to inject intermediate guidance without changing the target behavior. 
Recent analyses highlight its practical capabilities~\cite{wiewiora2003potential} in the field of multi-agent system~\cite{devlin2014potential, devlin2012dynamic} and robotics manipulation~\cite{camacho2021reward, chen2024boosting}. Process Reward Models (PRMs)~\cite{she2025r, zheng2025survey,li2024process, yin2025dynamic} can be viewed as a form of shaping that delivers step wise feedback in linguistic reasoning tasks~\cite{li2025codeprm, dai2024process, zhang2025lessons}. However, transferring such fine grained signals to object grounding is difficult, since intermediate reasoning steps is hard to reliably aligned with concrete image objects.
Rather than addressing this step-wise alignment challenge, our consolidation process draws on the core principle of reward shaping. It converts the improvements of the revision group over the initial response into dense feedback that shapes rewards and scales advantages, allowing the model to effectively learn from hard cases. 
\vspace{-5pt}

%% file: sec/3_method.tex
\input{images/workflow/workflow}
\section{Methodologies}
\subsection{Preliminary}
Let a sample from the dataset be $(x, q, y) \in \mathcal{D}$ , where each image \(x \in \mathcal{X}\) is paired with a question \(q \in \mathcal{Q}\). The corresponding supervision is \(y = (\mathbf{b}, \mathbf{p}) \in \mathcal{Y}\) , where \(\mathbf{b} \in \mathbb{R}^{N_x \times 4}\)  represents the bounding boxes of the referenced objects and \(\mathbf{p} \in \mathbb{R}^{N_x \times 2}\) provides point coordinates lying inside each object. Here, \(N_x\) is the number of annotated objects in image \(x\).
In the GRPO setting~\cite{shao2024grpo}, a policy model \(\pi_{\bm{\theta}}\) parameterised by \(\bm{\theta}\) is optimised using the rollouts collected by a previous (behaviour) policy snapshot \(\piold\), while a frozen reference policy \(\pi_{\textbf{\texttt{ref}}}\) provides a per-token forward KL regularisation term. For each training case, a response \(o \sim \piold(\cdot \mid x, q)\) is sampled.
Following~\cite{liu2025seg, liu2025visionreasoner}, the response is parsed as
\(o = (t, \hat{\mathbf{b}}, \hat{\mathbf{p}})\) using a deterministic parser, where \(t\) is the generated reasoning text,
\(\hat{\mathbf{b}} \in \mathbb{R}^{M\times 4}\) and \(\hat{\mathbf{p}} \in \mathbb{R}^{M\times 2}\) are the predicted
bounding boxes and points, respectively. Here \(M\) denotes the number of predicted objects and may
differ from \(N_x\).
These predicted boxes and points form the object-level grounding outputs that our optimisation aims to refine, identifying the referenced objects and their precise spatial locations.
To guide this optimisation, the reward is computed based on the grounding accuracy and the format of the response \(o\). Specifically, an accuracy term $\mathcal{R}_{\texttt{\textbf{acc}}}(o, y)$ measures spatial alignment with the ground truth (e.g., L1 or IoU), while a format term $\mathcal{R}_{\texttt{\textbf{format}}}(o)$ checks whether the response adheres to the expected output structure (e.g., \texttt{\small <think>...</think>} before \texttt{\small <answer>...</answer>})~\cite{liu2025seg, liu2025visionreasoner, huang2025samr1, cao2025groundr1}. 

\subsection{Group Revision and Policy Optimisation}
As shown in Fig.~\ref{fig:frame_work}, our training  starts with a sampling phase that generates an initial response, followed by a group of revised results conditioned on it. Next, a consolidation phase computes a shaping signal that quantifies the improvement achieved through the revision. Finally, this signal is used for both reward calculation and advantage scaling. \\
\textbf{Response Sampling.}  We first sample a \textit{single} response from the behavior policy as:
\vspace{-5pt}
\begin{equation}
(t\roundval{1}, \hat{\mathbf{b}}\roundval{1}, \hat{\mathbf{p}}\roundval{1}) = o\roundval{1} \sim \piold(\cdot \mid x, q),
\label{eq:1st_sample}
\vspace{-6.5pt}
\end{equation}
where \(t\roundval{1}\) is the reasoning text, and \(\hat{\mathbf{b}}\roundval{1}\), \(\hat{\mathbf{p}}\roundval{1}\) are the predicted bounding boxes and points\footnote{If the \(o\roundval{1}\) format is invalid or does not meet the expected specification, we fall back to the GRPO update procedure as described in~\cite{liu2025visionreasoner, liu2025seg}.}. 
This output defines the initial shaping set \(s\roundval{1}_{\mathrm{shape}} = \{\hat{\mathbf{b}}\roundval{1}, \hat{\mathbf{p}}\roundval{1}, y\}\).  \\
We then update the original question into a dialogue-style revised query
\(q\roundval{2} = \operatorname{\texttt{U}}(q, o\roundval{1})\)
via a fixed revision prompt template, incorporating cues from \(t\roundval{1}\) and spatial signals \((\hat{\mathbf{b}}\roundval{1}, \hat{\mathbf{p}}\roundval{1})\).
Based on the previous output, a \textit{group} of revision responses is sampled as:
\begin{equation}
o\roundval{2}_i \sim \piold\!\left(\cdot \mid x,\, q\roundval{2},\, o\roundval{1}\right), \quad \text{for } i=1,\dots,G.
\label{eq:2nd_round}
\end{equation}
Here, \(G\) denotes the group size, and each revision response \(o\roundval{2}_i\) serves as an alternative grounding hypothesis and potentially resolving ambiguities in challenging scenes. 
We then construct a per-candidate revision shaping set:
\begin{equation}
s\roundval{2}_{\mathrm{shape},i}
= \left\{\, o\roundval{2}_i,\, y  \;\middle|\; o\roundval{2}_i=(t\roundval{2}_i, \hat{\mathbf{b}}\roundval{2}_i, \hat{\mathbf{p}}\roundval{2}_i) \,\right\}.
\label{eq:collect_round2}
\end{equation}
For convenience, we denote the collected shaping sets uniformly as 
$s^{(r)}_{\mathrm{shape},i}$, where $r\in\{1,2\}$ 
and $i=1$ when $r=1$ (the single initial response) 
or $i\in\{1,\dots,G\}$ when $r=2$ (the group of revised outputs).
These shaping sets are used solely for computing rewards and advantage post-scaling.
\\
\textbf{Consolidation Process.} We compute the Hungarian matching between the predicted objects $\{(\hat{\mathbf{b}}_m\roundval{r}, \hat{\mathbf{p}}_m\roundval{r})\}_{m=1}^{M\roundval{r}_i}$ and the ground truth $\{(\mathbf{b}_n, \mathbf{p}_n)\}_{n=1}^{N_x}$, where $M\roundval{r}_i$ is the number of predictions and $N_x$ is the number of ground truth objects.
If $M\roundval{r}_i < N_x$, we pad the assignment with $N_x\!-\!M\roundval{r}_i$ dummy predictions (each incurring a unit cost) to penalise unmatched false negatives.
Thus, we obtain the matching set $\mathcal{A}_i\roundval{r} \subseteq \{1,\dots,M\roundval{r}_i\} \times \{1,\dots,N_x\}$.
Inspired by potential-based reward shaping~\cite{ng1999policy}, 
we define a potential function \(\Phi:s_{\mathrm{shape}}\roundval{r}\to\mathbb{R}\) that assigns a scalar to each set, enabling us to quantify the improvement introduced by the revision step.
Given the Hungarian matching \(\mathcal{A}_i\roundval{r}\) for \(s\roundval{r}_{\mathrm{shape},i}\), the alignment cost can be calculated as:
\begin{equation}
\Phi\!\left(s\roundval{r}_{\mathrm{shape}, i}\right)
\coloneqq \frac{1}{\lvert \mathcal{A}_i\roundval{r}\rvert}
\sum_{(m,n)\in \mathcal{A}_i\roundval{r}} e\roundval{r}_{m,n}.
\label{eq:alignment}
\vspace{-5pt}
\end{equation}
Here, \(e\roundval{r}_{m,n}\) denotes the pairwise cost between the object-level results and the ground truth, which is computed as:
\begin{equation}
\resizebox{.99\hsize}{!}{$
\begin{aligned}
e\roundval{r}_{m,n} =
\frac{1}{3}\left[
\left(1 - \mathrm{IoU}(\hat{\mathbf{b}}\roundval{r}_m , \mathbf{b}_n)\right)
+ \widetilde{\mathrm{L1}}(\hat{\mathbf{b}}\roundval{r}_m, \mathbf{b}_n)
+ \widetilde{\mathrm{L1}}(\hat{\mathbf{p}}\roundval{r}_m, \mathbf{p}_n)
\right],
\end{aligned}$}
\label{eq:pair-wise-measure}
\end{equation}
where \(\widetilde{\mathrm{L1}}\) denotes the L1 distance normalized by the image scale (for both boxes and points).
A \textit{lower} pairwise cost \(e\roundval{r}_{m,n}\) indicates \textit{better} grounding, as the predicted coordinates are closer to the ground truth.  \\ 
Based on the difference between the alignment costs of the two sets, we derive the shaping signal as:
\begin{equation}
\Delta \phi_{i} \;=\;
\max\!\left(0,\;
\frac{\Phi\big(s\roundval{1}_{\mathrm{shape}}\big)-\Phi\big(s\roundval{2}_{\mathrm{shape}, i}\big)}
{\Phi\big(s\roundval{1}_{\mathrm{shape}}\big)}
\right),
\label{eq:delta_phi}
\end{equation}
where $\Delta \phi_i>0$ means the revision improves upon the initial response and we avoid negative shaping signal in the optimisation.
The denominator $\Phi (s\roundval{1}_{\mathrm{shape}})$ provides a per-sample reference scale, 
so $\Delta \phi_{i}$ measures the relative improvement with respect to the initial response performance. 
When the initial is already close to the ground truth and the improvement space is limited (i.e., \(\Phi (s\roundval{1}_{\mathrm{shape}}) \approx 0\)), 
this scaling prevents the shaping signal from vanishing. \\
\textbf{Reward and Advantage.} We define the overall reward for each revision group candidate as:
\begin{equation}
    r_{i}
    = \mathcal{R}_{\texttt{\textbf{format}}}\!\left(o\roundval{2}_{i}\right)
    + \mathcal{R}_{\texttt{\textbf{acc}}}\!\left(o\roundval{2}_{i}, y\right)
    + \omega\,\Delta \phi_{i},
    \label{eq:reward}
\end{equation}
where \(\mathcal{R}_{\texttt{\textbf{format}}}\) and
\(\mathcal{R}_{\texttt{\textbf{acc}}}\) terms following~\cite{liu2025visionreasoner} to check the basic performance of \(o_i\roundval{2}\) for the format and the accuracy. The shaping signal \(\Delta \phi_i\) measures the improvement over the initial prediction, with \(\omega\) controlling its weight. \\
Following the GRPO optimisation~\cite{shao2024grpo},
we compute a z-score advantage
\(A_i=(r_i-\mu)/\sigma\), where \(\mu\) and \(\sigma\) are the mean and standard deviation computed over the \(G\) candidates within the same case.
We post-scale this advantage \(A_i\) to emphasise samples with larger improvements:
\begin{equation}
\hat{A}_{i} = (1 + \Delta \phi_i)\, A_{i},
\label{eq:adv}
\end{equation}
where \(\Delta\phi_i\) is the relative improvement defined in Eq.~\eqref{eq:delta_phi}. We highlight this post-scaling \(\hat{A}_i\) serves the different purpose with \(r_i\). The reward \(r_i\) determines preference and ranking in GRPO, whereas post-scaling from Eq.~\eqref{eq:adv} adjusts gradient magnitude, highlighting candidates with larger revision gains without altering the objective’s sign or form. \\
\textbf{Optimisation objective.} In the last, we formulate our  optimisation function as shown below:
\begin{equation}
\vspace{-10pt}
\small
\begin{split}
&\mathcal{J}_{\text{GRPO}}(\theta)
= \mathbb{E}\!\left[
\resizebox{.72\hsize}{!}{$
\begin{gathered}
 (x,q)\!\sim\!\mathcal{D},\; o\roundval{1}\!\sim\!\piold(\cdot\mid x,q),\!\\
 q\roundval{2}\!=\!\operatorname{\texttt{U}}(q,o\roundval{1}),\; \{o\roundval{2}_i\!\}_{i=1}^{G}\!\sim\!\piold(\cdot\mid x,q\roundval{2},o\roundval{1}\!)
\end{gathered}$}
\right] \\
&\quad \frac{1}{G}\sum_{i=1}^{G}
\min\!\Big(\ell_i(\bm{\theta})\,\hat A_i,\; \operatorname{clip}(\ell_i(\bm{\theta}),1-\epsilon,1+\epsilon)\hat A_i\Big)
-\beta \ \mathrm{KL}\big),
\end{split}
\end{equation}
where \scalebox{0.85}{$\ell_i(\bm{\theta}) \!\triangleq\! \dfrac{\pi_{\bm{\theta}}(o_i\roundval{2}\mid x, q\roundval{2})}{\piold(o_i\roundval{2}\mid x, q\roundval{2})}$} is the likelihood (importance) ratio between the current and  old policy. The ratio clipping with threshold $\epsilon$ limits the update induced by $\ell_i(\bm{\theta})$, preventing excessively large steps.
The KL term is $\mathrm{KL}\big(\pi_{\bm{\theta}}(\cdot\mid x,q\roundval{2})\,\|\,\pi_{\mathrm{\textbf{\texttt{ref}}}}(\cdot\mid x,q\roundval{2})\big)$, which regularises the policy towards a fixed reference with strength controlled by $\beta$.
The advantage $\hat{A}_i$ (from Eq.~\eqref{eq:adv}) is post-scaled by the shaping signal to improve sample efficiency during training.
\subsection{Inference}
Given an image–question pair $(x,q)$, we sample a single response $o=(t,\hat{\mathbf{b}},\hat{ \mathbf{p}})\sim \pi_{\bm{\theta}}(\cdot\mid x,q)$ with the current policy and parse the predicted objects $\{(\hat{\mathbf{b}}_m,\hat {\mathbf{p}}_m)\}_{m=1}^{M}$, where $M$ is the number of predicted objects. Here,  the counting result is reported based on $M$. Object localisation uses the predicted boxes $\{\hat{\mathbf{b}}_m\}_{m=1}^{M}$, and segmentation is obtained from the box–point pairs $\{(\hat{\mathbf{b}}_m,\hat{\mathbf{p}}_m)\}_{m=1}^{M}$ based on SAM2~\cite{ravi2024sam2}.

%% file: images/workflow/workflow.tex
\begin{figure*}[htbp!]
    \centering
    \includegraphics[width=\linewidth]{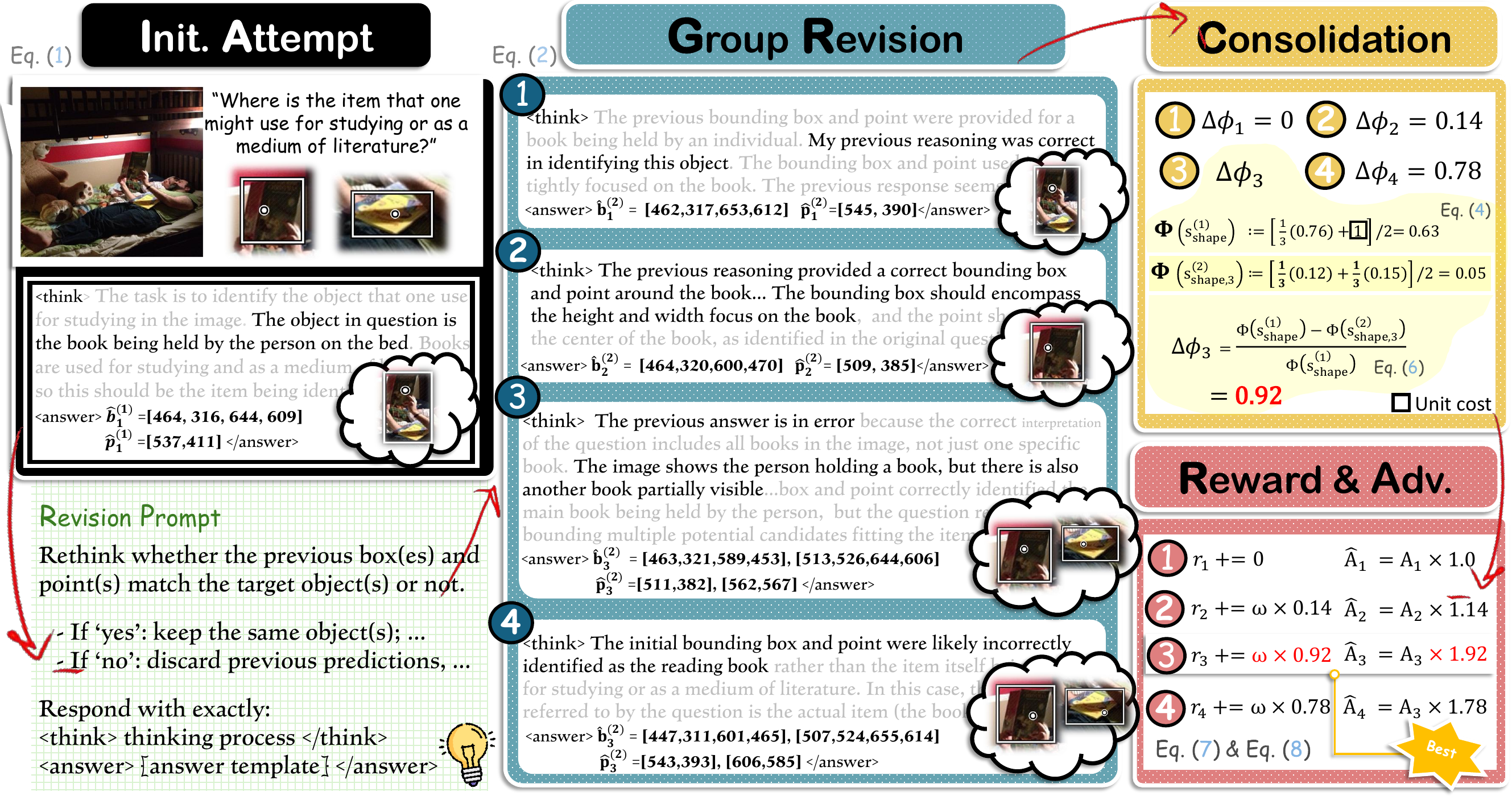}
\vspace{-15pt}\caption{\textbf{Illustration of our approach.} We first sample and collect the output from the initial response (Eq.~\eqref{eq:1st_sample}). A revision prompt is appended as a follow-up question to guide the sampling of a group of revised responses (Eq.~\eqref{eq:2nd_round}). We then collect the shaping sets from both results and compute the shaping signal \(\Delta \phi_i\) (Eq.~\eqref{eq:delta_phi}) between them. This signal is derived from the alignment cost (Eq.~\eqref{eq:alignment}), based on object-wise matching with the ground truth (Eq.~\eqref{eq:pair-wise-measure}). Finally, the signal is used for reward (Eq.~\eqref{eq:reward}) and advantage scaling (Eq.~\eqref{eq:adv}).}
    \label{fig:frame_work}
    \vspace{-10pt}
\end{figure*}

%% file: sec/4_experiment.tex
\section{Experiment}
\input{tables/main_experiment/main_result}
\input{tables/main_experiment/main_result_rec}
\textbf{Training Datasets.}
We conduct experiments under both single-~\cite{liu2025seg} and multi-object~\cite{liu2025visionreasoner} grounding setups.
For the \textbf{\textit{single-object}} setting, we adopt RefCOCOg~\cite{yu2016modeling} (modified following SegZero~\cite{liu2025seg}), which includes 9{,}000 image–question pairs, each referring to a specific object. All images are resized to $840\times 840$, and the ground truth consists of the one bounding box and two representative points inside the object.
For the \textbf{\textit{multi-object}} setting, we use the VisionReasoner7K dataset~\cite{liu2025visionreasoner}, which combines subsets of referring segmentation~\cite{yu2016modeling,gupta2019lvis,liu2023gres} and reasoning segmentation~\cite{yang2023lisa++} datasets, resulting in approximately 7{,}000 samples. To better reflect practical multi-object grounding scenarios, each sample may contain multiple objects, each annotated with a bounding box and a center point.\\
\textbf{Evaluation Dataset.}
We primarily focus on fine-grained grounding tasks, including segmentation and referring expression comprehension (REC), and additionally evaluate zero-shot generalisation to image-level tasks such as counting and visual question answering (VQA). For the \textbf{\textit{segmentation}} task, we evaluate on short, direct queries from RefCOCO (g/+)~\cite{yu2016modeling} and complex queries from ReasonSeg~\cite{lai2024lisa}, with ReasonSeg also serving as an out-of-domain benchmark for the \textbf{\textit{single}} setup. For the \textbf{\textit{REC}} task, we follow standard practice and convert predicted masks on these segmentation datasets~\cite{yu2016modeling,lai2024lisa} into bounding boxes, which are used to assess grounding accuracy.
For \textbf{\textit{counting}}, we evaluate on two benchmarks, CountBench~\cite{paiss2023teaching} and PixMoCount~\cite{deitke2025molmo}, by directly counting the number of predicted boxes without any additional counting supervision. We further evaluate our method on \textbf{\textit{VQA}} datasets~\cite{li2024seed,li2023evaluating,yue2025mmmu,fu2025mme,masry2022chartqa,liu2024ocrbench} to investigate how object-level Reinforcement learning influences image-level understanding.
\input{images/experiment/vqa/vqa}

\noindent\textbf{Implementation Details.}
Our method is build on top of Qwen2.5VL-7B~\cite{bai2025qwen2} as the base LVLM, with SAM2~\cite{ravi2024sam2} added for segmentation. The training pipeline adopts the VeRL~\cite{sheng2024hybridflow} framework with the vLLM~\cite{kwon2023efficient} engine, which supports both data and tensor parallelism. Following prior work~\cite{liu2025seg,liu2025visionreasoner}, we use a learning rate of $1\text{e-}6$ and weight decay of $0.01$. The default GRPO group size is 8, the batch size is 64, and the policy is updated every 2 steps. Please refer to Supp. Section \textcolor{red}{A} for more implementation details. \\
\noindent\textbf{Metrics.}
Following~\cite{liu2023gres,liu2025seg}, we use generalised Intersection over Union (gIoU) to evaluate pixel-level segmentation on ReasonSeg~\cite{lai2024lisa}, and cumulative IoU (cIoU) to assess referring segmentation~\cite{yu2016modeling}. We also report object-count accuracy and Acc@0.5 for REC, using an IoU threshold of 0.5 for predicted bounding boxes.
\input{tables/ablation_studies/ablation_main}
\subsection{Results across Different Tasks}
\textbf{Segmentation and Counting.}
In Tab.~\ref{tab:main_results}, we compare our method with prior SOTA approaches on both segmentation and counting. For segmentation, we observe consistent gains over other RFT-based methods~\cite{liu2025seg,liu2025visionreasoner,you2025segr1}. For example, on ReasonSeg~\cite{lai2024lisa} (test), our single-object setup achieves 61.11 gIoU, surpassing Seg-R1~\cite{you2025segr1} by 7.78 points. 
For counting, using the consistent checkpoint in multi-object setting, our method attains an average accuracy of 79.97\% across two datasets~\cite{deitke2025molmo, paiss2023teaching}, and we observe a 5.64 \% improvement over VisionReasoner~\cite{liu2025visionreasoner}. Benefiting from the group-revision paradigm, these results indicate more accurate object grounding and localization (both boxes and points) across diverse and challenging scenarios.\\
\noindent\textbf{Referring Expression Comprehension.}
As shown in Tab.~\ref{tab:rec_results}, our  model (multi) achieves the best overall Acc@0.5 of 85.80\%, surpassing the previous SOTA VisionReasoner~\cite{liu2025visionreasoner} by 1.0 points. Although our method shows lower performance on RefCOCO+ (testB) than VisionReasoner, this is largely due to its fewer false-positive boxes: VisionReasoner produces many extra boxes, which inflate Acc@0.5 but are consistent with its lower segmentation quality (shown in Tab.~\ref{tab:main_results}).
Compared with Seg-Zero~\cite{liu2025seg} under the single-object setting, our method yields notable improvements of 4.58\% and 1.81\% on ReasonG (val and test, boxes converted from ReasonSeg~\cite{lai2024lisa}). These results suggest the benefit of our group-revision optimisation paradigm for improving visually grounded bounding box localisation. \\
\noindent\textbf{General VQA.}
In Tab.~\ref{tab:vqa}, we use the \texttt{\small llms-eval}~\cite{zhang2024lmmsevalrealitycheckevaluation} toolkit to assess performance across diverse VQA benchmarks. Our method (multi) shows consistent gains over~\cite{bai2025qwen2}, achieving +0.55\% on SeedBenchV2+~\cite{li2024seed} and +1.25\% (average) on MMU-pro~\cite{yue2024mmmu}. These results suggest that object-level grounding improves instance-level localisation and benefits overall image-level understanding.
\input{tables/ablation_studies/potential_weight}
\input{images/experiment/weight/post_ablation}
\subsection{Ablation Studies}
\textbf{Ablation in our main contribution.} 
As shown in Tab.~\ref{tab:main_ablation}, both revision sampling and consolidation with reward shaping provide consistent gains in single and multi setups. Training with the revision group steadily improves over directly generated answers (as in GRPO). In the single-object setting, performance on ReasonSeg (val)~\cite{lai2024lisa} improves from 62.54\% to 64.97\%, and on RefCOCOg (val)~\cite{yu2016modeling} from 70.84\% to 70.92\%. Consolidation yields larger boosts by up-weighting higher-reward candidates and amplifying gradients, further increasing these results to 66.99\% and 73.84\%. The effect is most pronounced on counting: PixMoCount (val)~\cite{deitke2025molmo} rises from 70.84\% to 73.20\% with revision, and to 75.89\% with consolidation. Overall, revision sampling introduces diverse hypotheses and recovers learning signals, while consolidation effectively amplifies these benefits, leading to robust performance gains.\\
\noindent\textbf{Ablation on reward weight (\(\omega\)).}
In Tab.~\ref{tab:omega_weight}, we study the effect of the shaping weight in the reward on segmentation~\cite{lai2024lisa} and REC~\cite{yu2016modeling} under the single setup. We observe that \(\omega = 3\) and \(7\) yields slightly better results on specific splits, whereas \(\omega=5\) achieves the best overall results.

\input{tables/ablation_studies/potential_component}
 \input{images/experiment/delta_phi_pdf/pdf_fig}
\noindent\textbf{Ablation on advantage post-scaling.}
We evaluate the impact of advantage post-scaling from Eq.~\eqref{eq:adv} in Fig.~\ref{fig:post_ablation}. Consistent gains are observed on all four datasets for both validation and test splits, including improvements of 0.53\% and 1.71\% points on ReasonSeg~\cite{lai2024lisa}, and 1.86\% and 0.97\% points on PixMoCount~\cite{deitke2025molmo}. These results indicate that scaling strengthens gradient updates during consolidation.
 \\
\noindent\textbf{Consolidation signal distribution.}
Fig.~\ref{fig:pdf} shows the probability density function (PDF) of \(\Delta\phi\) from Eq.~\eqref{eq:delta_phi} for both single and multi setups. This PDF illustrates how consolidation gains contribute to the reward term and to advantage scaling.
In the single-object setting, the group-wise \textcolor{steelblue!80}{mean} is concentrated around 0-0.2, indicating mild but consistent improvements per sample, while the \textcolor{orange!80}{max} exhibits a heavy right tail with most mass between 0.8 and 1.0. This pattern suggests that strong revision candidates do exist and can be effectively up-weighted, while most cases remain stable. Crucially, the distribution indicates that initial responses are generally of competitive quality and are not intentionally degraded to inflate \(\Delta\phi\), reducing the risk of \textbf{\textit{reward hacking}}. Together, these observations support the stability and integrity of our group-revision optimisation paradigm.
\input{tables/ablation_studies/hallucination}

\noindent\textbf{Reasoning Process Evaluation.} As shown in Tab.~\ref{tab:failure_types}, we assess CoT reasoning hallucinations based on GPT-5. We consider three types: \textit{off-topic} (reasoning does not follow the query), \textit{attribute mismatch} (the description does not correctly describe the object inside the bounding box), and \textit{vague claim} (the description is overly generic and not specific to the object). Our method (multi) produces fewer hallucinations than VisionReasoner~\cite{liu2025visionreasoner}, including -52 cases of attribute mismatch and -22 cases of vague claims. While the single-object setup reduces vague claims, it causes more mismatches, whereas multi-object revision offers more consistent grounding in complex scenes.
The fewer hallucination cases in our methods show that optimising on hard cases helps the model produce more accurate CoT.
\subsection{Visualisation}
\begin{figure}[t!]
    \centering
    \includegraphics[width=\linewidth]{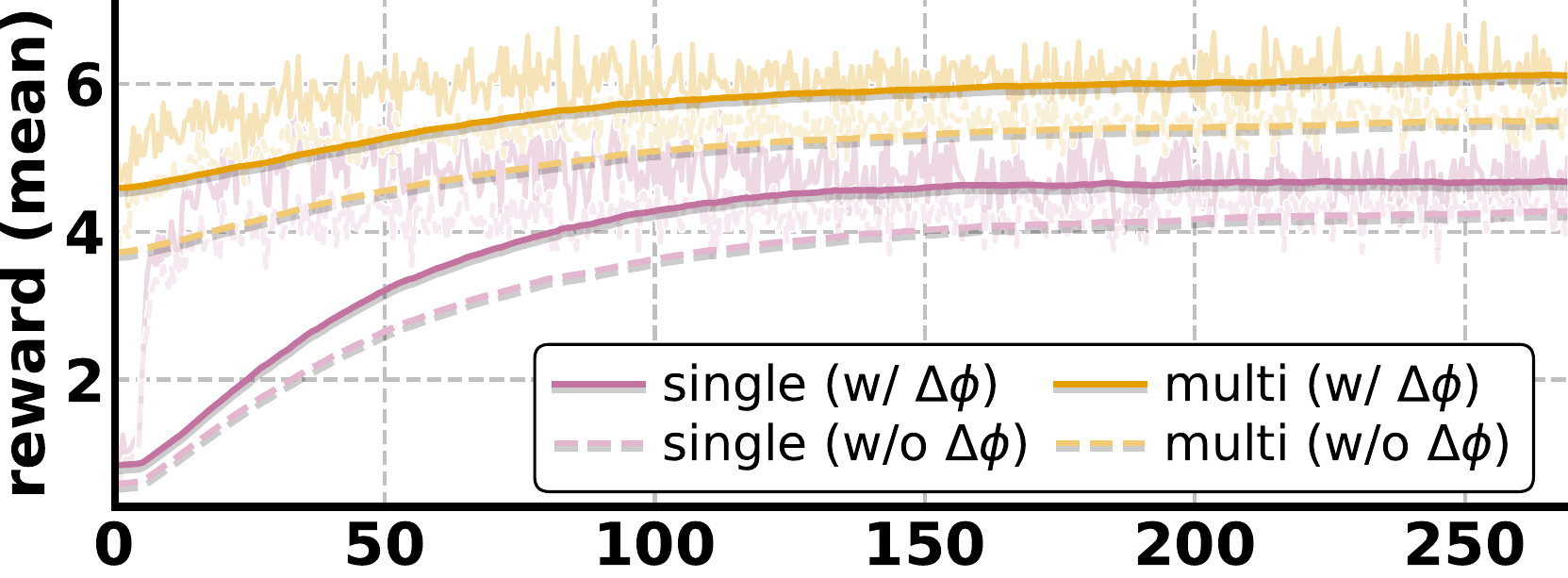}
    \vspace{-20pt}
\caption{\textbf{Reward curves} over training steps for both single- and multi-object setups. Each line shows the mean reward of the revision group in condition of with and without the \(\Delta\phi\) term.}
    \label{fig:reward_curve}
    \vspace{-15pt}
\end{figure}
\textbf{Reward Curves.} Fig.~\ref{fig:reward_curve} presents the reward trajectories for both single- and multi-object setups, comparing our method against GRPO (without the revision process). The group revision responses consistently achieves higher rewards, suggesting that revision gains (especially from hard samples) are effectively transformed into stronger learning signals.

\begin{figure}[t!]
    \centering
    \includegraphics[width=\linewidth]{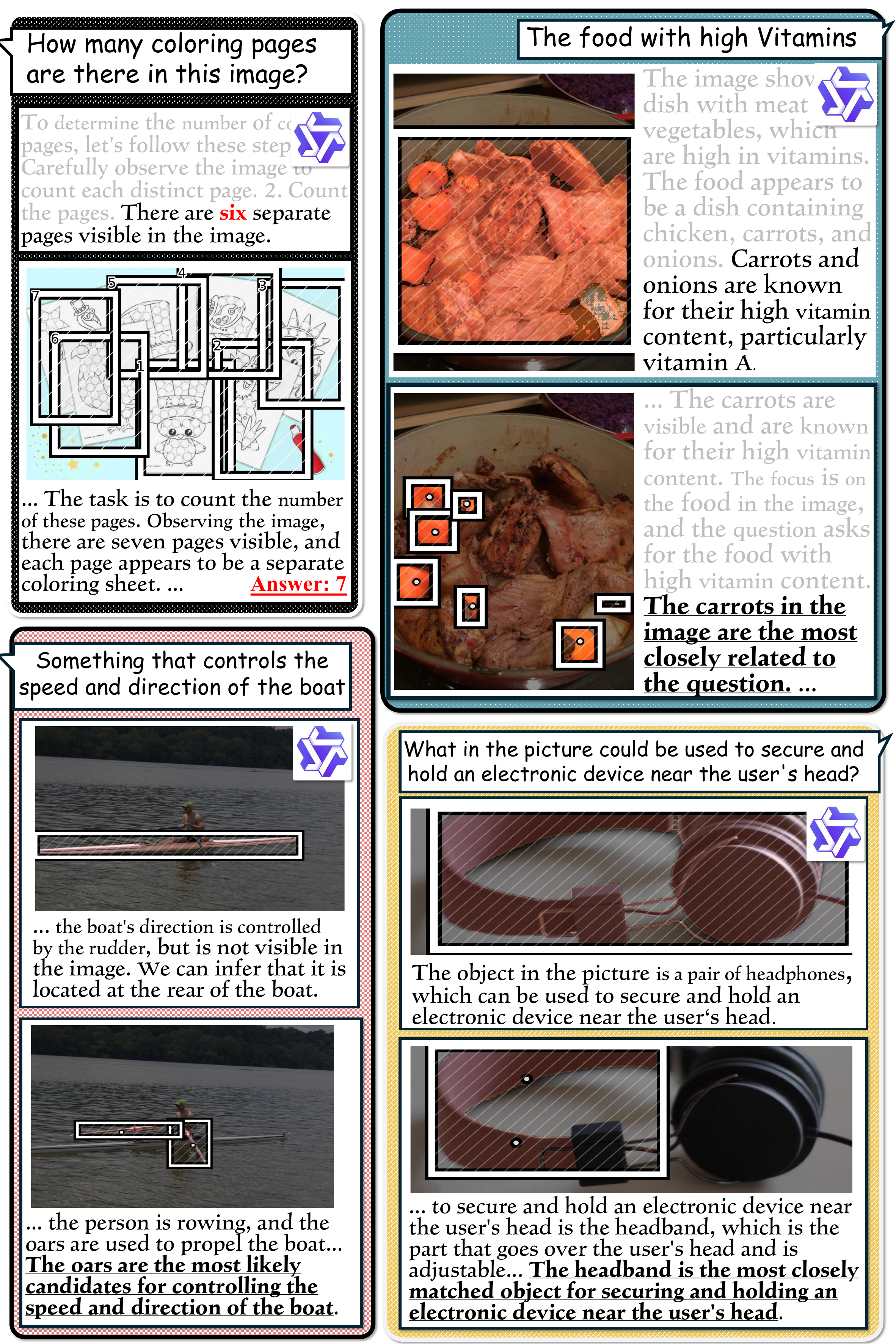}
    \vspace{-20pt}
    \caption{\textbf{Qualitative comparisons} on the CountingBench~\cite{paiss2023teaching} and ReasoningSeg~\cite{lai2024lisa} datasets. For each example, we present the outputs from Qwen2.5VL-7B~\cite{bai2025qwen2} (top row) and our method (bottom), including both CoT reasoning and spatial grounding.}
    \label{fig:quality}
    \vspace{-15pt}
\end{figure}
\noindent \textbf{Qualitative Results.}
In Fig.~\ref{fig:quality}, we compare our method with Qwen2.5VL-7B~\cite{bai2025qwen2} on both the reasoning segmentation~\cite{lai2024lisa} and counting~\cite{paiss2023teaching} benchmarks. Our approach shows consistently better visual performance. For example, in the case of the query `\texttt{\small Something that controls the speed and direction of the boat}', our model produces more coherent reasoning (e.g., correctly identifying the oars) and delivers more accurate spatial grounding.

%% file: tables/main_experiment/main_result.tex
\begin{table*}[t!]
\renewcommand{\arraystretch}{1.1}
\caption{
\textbf{Comparison with SoTA on Segmentation and Counting Tasks.}  
We compare our method with prior SoTA approaches in reasoning segmentation~\cite{lai2024lisa} (evaluated with gIoU), referring segmentation~\cite{yu2016modeling} (cIoU), and counting~\cite{paiss2023teaching, deitke2025molmo} (accuracy).  
\(*\) denotes zero-shot performance on the reasoning segmentation task;  
\(\#\) indicates re-implemented results based on officially released checkpoints;  
\({\dagger}\) marks methods using a consistent checkpoint across all benchmarks.  
The best results are shown in \textcolor{darkred}{red}, and the second-best are \ul{underlined}.  
We highlight these {\setlength{\fboxsep}{0pt}\colorbox{gray!10}{\strut SFT methods}} generally use larger training sets than RFT methods, please see Supp. Section \textcolor{red}{B} for more details.
}
\label{tab:main_results}
\vspace{-8pt}
\centering
\resizebox{\linewidth}{!}{
\begin{tabular}{?r?ccccccc?c?ccc?c?}
\specialrule{1.2pt}{0pt}{0pt}
\multicolumn{1}{?c?}{\multirow{3}{*}{Method}} & \multicolumn{7}{c?}{Segmentation (gIoU \& cIoU)}                                & \multicolumn{1}{c?}{\multirow{3}{*}{Average}} & \multicolumn{3}{c?}{Counting (Acc)}        & \multicolumn{1}{c?}{\multirow{3}{*}{Average}} \\
\cline{2-8}\cline{10-12}
\multicolumn{1}{?c?}{}                        & \multicolumn{2}{c}{ReasonSeg} & \multicolumn{2}{c}{RefCOCO} & \multicolumn{2}{c}{RefCOCO+} & RefCOCOg & \multicolumn{1}{c?}{}                         & \multicolumn{2}{c}{Pixmo} & Count   & \multicolumn{1}{c?}{}                         \\
\multicolumn{1}{?c?}{}                        & val            & test           & testA    & testB    & testA      & testB     & test     & \multicolumn{1}{c?}{}                         & val          & test       & test    & \multicolumn{1}{c?}{}                         \\
\specialrule{1.2pt}{0pt}{0pt}
\rowcolor{gray!10}
LLaVA-OV-7B~\cite{li2024llava}~{\scriptsize \textcolor{gray}{[arxiv 2024]}}                                 & -              & -              & 58.1    & --       & 47.1     & --        & 55.6     & -                                            & -            & -          & -       & -                                            \\             
\rowcolor{gray!10}
LISA-7B~\cite{lai2024lisa}~{\scriptsize \textcolor{gray}{[CVPR 2024]}}                                     & 44.4           & 36.8           & 79.1    & 72.3      &  70.8      & 58.1        & 70.6      &                                         & -            & -          & -       & -                                            \\        
\rowcolor{gray!10}
PixelLM-7B~\cite{ren2024pixellm}~{\scriptsize \textcolor{gray}{[CVPR 2024]}}                               & -              & -              & 76.5    & 68.2       & 71.7     & 58.3        & 70.5     & -                                            & -            & -          & -       & -                                            \\     
\rowcolor{gray!10}
PerceptionGPT-7B~\cite{pi2024perceptiongpt}~{\scriptsize \textcolor{gray}{[CVPR 2024]}}                    & -              & -              & 78.6    & 71.7       & 73.9     & 61.3         & 71.7     & -                                            & -            & -          & -       & -                                            \\        
\rowcolor{gray!10}
SEGLLM~\cite{wang2024segllm}~{\scriptsize \textcolor{gray}{[ICLR 2025]}}                                   & 57.2           & 52.4           & \textcolor{darkred}{81.5}    & \textcolor{darkred}{75.4}       & 73.0     & \ul{62.5}        & \ul{73.6}     & 67.9                                            & -            & -          & -       & -                                            \\      
\rowcolor{gray!10}
Read-7B~\cite{qian2024reasoning}~{\scriptsize \textcolor{gray}{[CVPR 2025]}}                                & 59.8           & 56.8           & 80.2    &  \ul{73.2}       & 73.7     & 60.4        & 71.4     & 67.9                                            & -            & -          & -       & -                                            \\    
Qwen2-VL-7B\(^{*\#\dagger}\)~\cite{wang2024qwen2}~{\scriptsize \textcolor{gray}{[arxiv 2024]}}                               & 44.5           & 38.7           & 58.9 & 48.0       & 52.3 & 41.9        & 52.1 &   48.1                                         & 29.9         & 48.0       & 76.5    & 51.5                                         \\
Qwen2.5-VL-7B\(^{*\#\dagger}\)~\cite{bai2025qwen2}~{\scriptsize \textcolor{gray}{[arxiv 2025]}}                             & 56.9           & 52.1           & 77.9 & 66.5       & 74.0 & 55.6         & 70.9 &  64.8                                          & 63.3         & 67.9       & 76.0    & 69.1                                         \\
Seg-R1-7B\(^{*}\)~\cite{you2025segr1}~{\scriptsize \textcolor{gray}{[arXiv 2025]}}                         & 58.6           & 56.7           & 78.7     & 67.6       & 70.9     & 57.9        & 71.4     & 66.0& -            & -          & -       & -                                            \\
Seg-Zero-7B\(^{*}\)~\cite{liu2025seg}~{\scriptsize \textcolor{gray}{[arXiv 2025]}}                         & 62.6           & 57.5           & 80.3     & 72.2$^\#$       & \ul{76.2}     & 62.3$^\#$        & 72.6     & 69.1                                     & -            & -          & -       & -                                            \\
VisionReasoner-7B\(^{\dagger}\)~\cite{liu2025visionreasoner}~{\scriptsize \textcolor{gray}{[arXiv 2025]}}   & 66.3           & \ul{63.6}           & 77.4$^\#$ & 67.6$^\#$      & 71.1$^\#$ & 55.8$^\#$        & 68.3$^\#$ &          67.2                                  & 70.1         & 69.5       & 87.6    & 75.7                                         \\           

\rowcolor{cyan!8}    
Ours (single-object)\(^{*}\)                                                                                & \ul{66.99}          & 61.11          & \ul{80.79}    & \ul{73.16}       & \textcolor{darkred}{76.72}    & \textcolor{darkred}{62.92}        & \textcolor{darkred}{74.26}    &      \textcolor{darkred}{70.85}                                       & -            & -          & -       & -                                            \\  
\rowcolor{cyan!10}
Ours (multi-object)\(^{\dagger}\)                                                                            & \textcolor{darkred}{67.48}          & \textcolor{darkred}{66.73}          & 78.04     & 69.53       & 73.30     & 59.34        & 71.08     &   \ul{69.36}                                         & \textcolor{darkred}{75.89}        & \textcolor{darkred}{72.97}      & \textcolor{darkred}{91.04}   & \textcolor{darkred}{79.97}                                        \\               
 
\specialrule{1.2pt}{0pt}{0pt}        
\end{tabular}}
\vspace{-8pt}
\end{table*}

%% file: tables/main_experiment/main_result_rec.tex
\begin{table*}[]
\caption{
\textbf{Comparison with SoTA methods on the REC task.}  
We compare with SoTA methods on ReasonG (boxes converted from reasoning segmentation~\cite{lai2024lisa}) and referring grounding (RefCOCO (+/g)~\cite{yu2016modeling}).
\(\#\) denotes re-implementations using official checkpoints; \({\dagger}\) denotes a single checkpoint across all benchmarks.
The best results are shown in \textcolor{darkred}{red}, and the second-best are \ul{underlined}.
}
\vspace{-8pt}
\label{tab:rec_results}
\centering
\resizebox{.93\linewidth}{!}{%
\begin{tabular}{?r?cc?ccc?ccc?cc?c?}
\specialrule{1.2pt}{0pt}{0pt}
\renewcommand{\arraystretch}{1.2}
\multirow{3}{*}{\makebox[-120pt][c]{Method}}   & \multicolumn{10}{c?}{Referring Expression Comprehension (Acc@0.5)} & \multicolumn{1}{c?}{\multirow{3}{*}{Average}} \\
\cline{2-11}
& \multicolumn{2}{c?}{ReasonG} & \multicolumn{3}{c?}{RefCOCO} & \multicolumn{3}{c?}{RefCOCO+} & \multicolumn{2}{c?}{RefCOCOg} & \\
& val & test & val & testA & testB & val & testA & testB & val & test & \\
\specialrule{1.2pt}{0pt}{0pt}
\rowcolor{gray!10}
PerceptionGPT-7B~\cite{pi2024perceptiongpt}~{\scriptsize \textcolor{gray}{[CVPR 2024]}} &  -- &  -- & 88.6 & \textcolor{darkred}{92.5} & 84.6   & 82.1 & 88.6 & 74.2  & 84.1 & 85.2 & -- \\
\rowcolor{gray!10}
VistaLLM-7B~\cite{pramanick2024jack}~{\scriptsize \textcolor{gray}{[ECCV 2024]}} &--&--&  88.1 & 91.5 & 83.0 & 82.9 & \textcolor{darkred}{89.8} & 74.8 & 83.6 & 84.4 &-- \\
\rowcolor{gray!10}
Elysium-7B~\cite{wang2024elysium}~{\scriptsize \textcolor{gray}{[ECCV 2024]}} &--&--& 89.1 & 92.1 & 85.0 & 82.9 & 88.9 & 75.6 & 82.9 & 83.6& --\\
\rowcolor{gray!10}
Groma-7B~\cite{ma2024groma}~{\scriptsize \textcolor{gray}{[ECCV 2024]}} &--&--& \ul{89.5} & 92.1 & \ul{86.3} & 83.9 & 88.9 & \ul{78.1} & 86.3 & 87.0 &-- \\
Qwen2-VL-7B\(^{*\#\dagger}\)~\cite{wang2024qwen2}~{\scriptsize \textcolor{gray}{[arxiv 2024]}}   & 54.6 & 46.9 & 80.8 & 83.9 & 58.9   & 72.5 & 76.5 & 54.5   & 77.3 & 78.2 & 68.4 \\
Qwen2.5-VL-7B\(^{*\#\dagger}\)~\cite{bai2025qwen2}~{\scriptsize \textcolor{gray}{[arxiv 2025]}} & 68.9 & 59.8 & 88.8 & 91.7 & 81.4   & 82.3 & 88.2 & 69.2   & 84.7 & 85.7 & 80.1 \\
SegZero-7B\(^{*\#}\)~\cite{liu2025seg}~{\scriptsize \textcolor{gray}{[arxiv 2025]}}    & 69.3 & 64.6 & 89.3 & 91.5 & 81.9   & 82.0 & 87.6 & 74.7   & 86.1 & 86.3 & 81.3 \\
VisionReasoner-7B\(^\dagger\)~\cite{liu2025visionreasoner}~{\scriptsize \textcolor{gray}{[arxiv 2025]}} & \ul{80.1}\(^{\#}\)& \ul{78.5}\(^{\#}\) & 88.6 & 90.6 & 84.7\(^{\#}\)   & 83.6 & 87.9 & \textcolor{darkred}{80.2}\(^{\#}\)   & 86.1 & 87.5 & \ul{84.8} \\
\rowcolor{cyan!10}
Ours (single-object) & 73.98 & 66.41 & \textcolor{darkred}{90.72} & \ul{92.36} & \textcolor{darkred}{86.67}   & \textcolor{darkred}{84.49} & \ul{89.28} & 74.26   & \ul{87.64} & \ul{87.97} & 83.38 \\
\rowcolor{cyan!10}
Ours (multi-object)\(^\dagger\) & \textcolor{darkred}{83.67} & \textcolor{darkred}{81.20} & \ul{89.49} & 91.55 & 85.05   & \ul{84.37} & 88.33 &  77.60  & \textcolor{darkred}{88.44} & \textcolor{darkred}{88.32} & \textcolor{darkred}{85.80} \\
\specialrule{1.2pt}{0pt}{0pt}
\end{tabular}
}
\vspace{-10pt}
\end{table*}

%% file: images/experiment/vqa/vqa.tex
\begin{table}[t]
\renewcommand{\arraystretch}{1.1}
\centering
\caption{\textbf{General VQA.} Comparison on diverse VQA benchmarks for general image understanding. Best results are shown in \textcolor{darkred}{red}.}
\label{tab:vqa}
\vspace{-8pt}

\begin{subtable}{\linewidth}
\centering
\resizebox{\linewidth}{!}{%
\begin{tabular}{!{\vrule width 1.5 pt}l|c|c|c!{\vrule width 1.5pt}}
\specialrule{1.5pt}{0pt}{0pt}
\multicolumn{1}{!{\vrule width 1.5pt}c|}{Model} 
& SeedBenchV2+~\cite{li2024seed} & ChartQA~\cite{masry2022chartqa} & OCRBench~\cite{liu2024ocrbench} \\
\cline{2-4}
\multicolumn{1}{!{\vrule width 1.5pt}c|}{}
& Acc. & Acc. & Num. \\
\specialrule{1.5pt}{0pt}{0pt}
Qwen2.5VL-7B~\cite{bai2025qwen2} 
& 70.4  & 83.6  & 849 \\
Ours (single) 
& 70.75 & 84.03 & 853 \\
Ours (multi) 
& \textcolor{darkred}{70.95} & \textcolor{darkred}{84.96} & \textcolor{darkred}{861} \\
\specialrule{1.5pt}{0pt}{0pt}
\end{tabular}}
\end{subtable}

\vspace{2pt}

\begin{subtable}{\linewidth}
\centering
\resizebox{\linewidth}{!}{%
\begin{tabular}{!{\vrule width 1.5 pt}l|c|cc|cc!{\vrule width 1.5pt}}
\specialrule{1.5pt}{0pt}{0pt}
\multicolumn{1}{!{\vrule width 1.5pt}c|}{\multirow{2}{*}{Model}} 
& \multirow{2}{*}{POPE~\cite{li2023evaluating}} 
& \multicolumn{2}{c|}{MMMU-Pro~\cite{yue2025mmmu}} 
& \multicolumn{2}{c!{\vrule width 1.5pt}}{MME~\cite{fu2025mme}} \\
\cline{3-6}
\multicolumn{1}{!{\vrule width 1.5pt}c|}{} 
&  
& Vis & Std 
& Cog. & Percept. \\
\specialrule{1.5pt}{0pt}{0pt}
Qwen2.5VL-7B~\cite{bai2025qwen2}     
& 87.1  & 32.5  & 36.4  & 619.0 & 1691.9     \\
Ours (single) 
& \textcolor{darkred}{88.24} & 33.12 & 37.11 & 627.14 & 1698.11 \\
Ours (multi) 
& 88.21 & \textcolor{darkred}{33.70} & \textcolor{darkred}{37.75} & \textcolor{darkred}{630.00} & \textcolor{darkred}{1708.01} \\
\specialrule{1.5pt}{0pt}{0pt}
\end{tabular}}
\end{subtable}
\vspace{-1pt}
\end{table}

%% file: tables/ablation_studies/ablation_main.tex
\begin{table*}[]
\caption{\textbf{Ablation study of our approach} on both single-object and multi-object setups. We apply our method on ReasonSeg~\cite{lai2024lisa} and RefCOCO+~\cite{yu2016modeling} for the segmentation task, RefCOCOg~\cite{yu2016modeling} for the REC task, and PixMoCount~\cite{deitke2025molmo} for counting (multi-object only). The w/ revision refers to Eq.~\eqref{eq:2nd_round}, and w/ consolidation refers to shaping signal in both Eq.~\eqref{eq:reward} and~\eqref{eq:adv}. The best results are shown in \textcolor{darkred}{red}.}
\vspace{-10pt}
\label{tab:main_ablation}
\resizebox{\linewidth}{!}{
\begin{tabular}{?r?cccc|cc?cccc|cc|cc?}
\specialrule{1.2pt}{0pt}{0pt}
\multicolumn{1}{?c?}{\multirow{4}{*}{Ablation Study}} & \multicolumn{6}{c?}{Single-Object} & \multicolumn{8}{c?}{Multi-Object} \\
\cline{2-15}
                                & \multicolumn{4}{c|}{Segmentation} & \multicolumn{2}{c?}{REC} & \multicolumn{4}{c|}{Segmentation} & \multicolumn{2}{c|}{REC} & \multicolumn{2}{c?}{Counting} \\
\cline{2-15}
                                & \multicolumn{2}{c}{ReasonSeg} & \multicolumn{2}{c|}{RefCOCOg} & \multicolumn{2}{c?}{RefCOCOg} & \multicolumn{2}{c}{ReasonSeg} & \multicolumn{2}{c|}{RefCOCOg} & \multicolumn{2}{c|}{RefCOCOg} & \multicolumn{2}{c?}{Pixmo} \\
                                & val & test & val & testA & val & testA & val & test & val & testA & val & testA & val & test \\
\specialrule{1.2pt}{0pt}{0pt}
baseline                        &  62.54    & 58.67     &  70.84   &  72.17     &  85.92   &   86.67  &  65.37    &   64.23    &  66.87  &  68.92     & 87.26   &  87.20     &  70.84   &  70.53    \\
w/ revision                        &   64.97  &  60.24    &  70.92  &     73.80  &   86.79  & 87.03      &   66.54  &    65.91  &  69.92   & 70.04      &  88.13  &  \textcolor{darkred}{88.54}   &  73.20   & 71.94     \\
w/ consolidation                        & \textcolor{darkred}{66.99}    & \textcolor{darkred}{61.11}     &  \textcolor{darkred}{73.58}  &  \textcolor{darkred}{74.26}     &   \textcolor{darkred}{87.64}  & \textcolor{darkred}{87.97}      & \textcolor{darkred}{67.48}    & \textcolor{darkred}{66.73}     &  \textcolor{darkred}{70.67}   & \textcolor{darkred}{71.08}      &  \textcolor{darkred}{88.32}   &  88.44     &   \textcolor{darkred}{75.89}  &   \textcolor{darkred}{72.97}   \\

\specialrule{1.2pt}{0pt}{0pt}
\end{tabular}}
\vspace{-15pt}
\end{table*}

%% file: tables/ablation_studies/potential_weight.tex
\begin{table}[t!]
\centering
\caption{\textbf{Ablation of Consolidation Weight.} We vary the parameter $\omega$ from Eq.~\eqref{eq:reward} under the single-object setting and assess its impact on segmentation~\cite{lai2024lisa,yu2016modeling} and REC~\cite{yu2016modeling} tasks.}
\label{tab:omega_weight}
\vspace{-9.4pt}
\resizebox{\linewidth}{!}{%
\begin{tabular}{?c?cc|cc|cc?}
\specialrule{1.2pt}{0pt}{0pt}
\multirow{3}{*}{weight (\(\omega\))} & \multicolumn{4}{c|}{Segmentation} & \multicolumn{2}{c?}{REC} \\
\cline{2-7}
 & \multicolumn{2}{c}{ReasonSeg} & \multicolumn{2}{c|}{RefCOCOg } & \multicolumn{2}{c?}{RefCOCOg} \\
\cline{2-7}
 & val & test & val & testA & val & testA \\
\specialrule{1.2pt}{0pt}{0pt}
3 & 64.36 & 59.67 & 72.86 & 73.63 & \textcolor{darkred}{87.72} & 87.89 \\
\textcolor{darkred}{5} & \textcolor{darkred}{66.99} & 61.11 & \textcolor{darkred}{73.58} & \textcolor{darkred}{74.26} & 87.64 & \textcolor{darkred}{87.97} \\
7 & 65.84 & \textcolor{darkred}{62.07 }& 72.73 & 73.92 & 87.03 & 87.14 \\
\specialrule{1.2pt}{0pt}{0pt}
\end{tabular}%
}
\end{table}

%% file: images/experiment/weight/post_ablation.tex
\begin{figure}[t!]
    \centering
    \includegraphics[width=\linewidth]{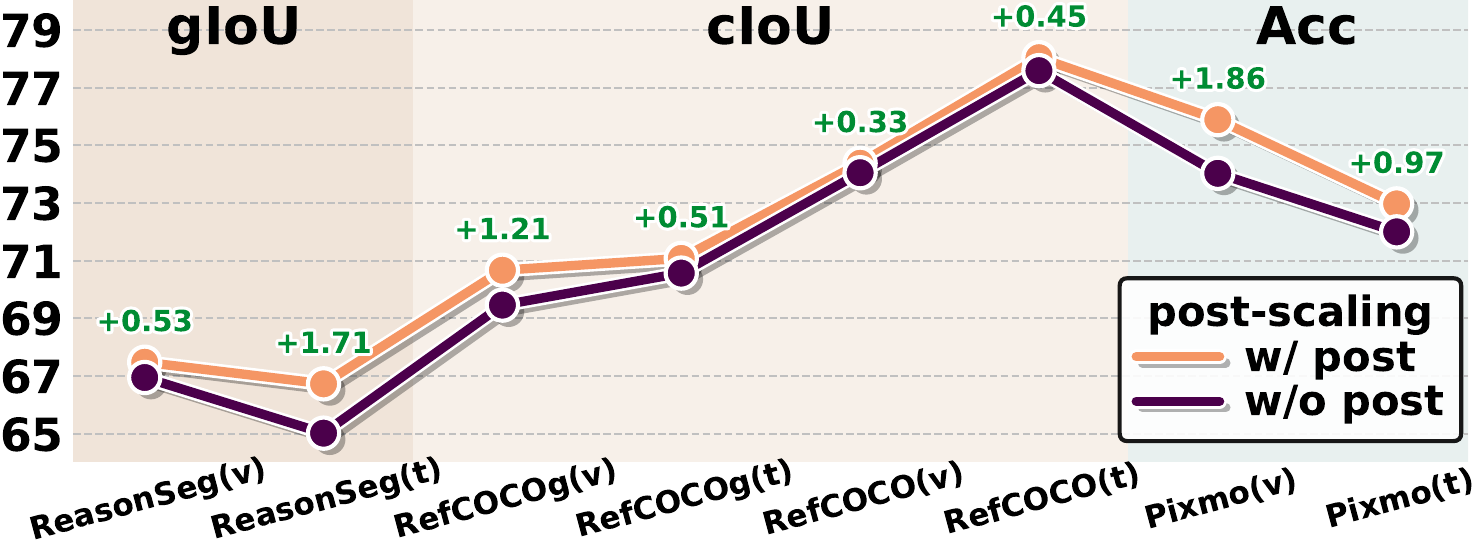}
    \vspace{-15pt}
\caption{\textbf{Ablation of Post Scaling}. We evaluate our method with the effectiveness of advantage post-scaling from Eq.~\eqref{eq:adv} in the segmentation~\cite{lai2024lisa,yu2016modeling} and counting tasks~\cite{deitke2025molmo} in single-object setting. }
    \label{fig:post_ablation}
    \vspace{-15pt}
\end{figure}

%% file: tables/ablation_studies/potential_component.tex
\begin{table}[t!]
\centering
\caption{\textbf{Ablation of Consolidation Components.} We assess the effect of removing point or box shaping signals (from Eq.~\eqref{eq:pair-wise-measure}) on segmentation~\cite{lai2024lisa, yu2016modeling} and REC~\cite{yu2016modeling} tasks.}
\vspace{-9.4pt}
\resizebox{\linewidth}{!}{%
\begin{tabular}{?r?cc|cc|cc?}
\specialrule{1.2pt}{0pt}{0pt}
\multirow{3}{*}{Variant} & \multicolumn{4}{c|}{Segmentation} & \multicolumn{2}{c?}{REC} \\
\cline{2-7}
 & \multicolumn{2}{c}{ReasonSeg} & \multicolumn{2}{c|}{RefCOCOg} & \multicolumn{2}{c?}{RefCOCOg} \\
\cline{2-7}
 & val & test & val & testA & val & testA \\
\specialrule{1.2pt}{0pt}{0pt}
w/ both   & \textcolor{darkred}{66.99} & \textcolor{darkred}{61.11}  & \textcolor{darkred}{73.58}  & \textcolor{darkred}{74.26}  & \textcolor{darkred}{87.64}  & \textcolor{darkred}{87.97}  \\
w/o box   & 65.32  & 60.28 & 71.06 & 73.48 & 86.04  & 86.58 \\
w/o point & 65.90 &  59.97 & 72.64  & 73.52 & 87.59 & 87.76 \\
\specialrule{1.2pt}{0pt}{0pt}
\end{tabular}%
}
\vspace{-10pt}
\end{table}

%% file: images/experiment/delta_phi_pdf/pdf_fig.tex
\begin{figure}
    \centering
\includegraphics[width=\linewidth]{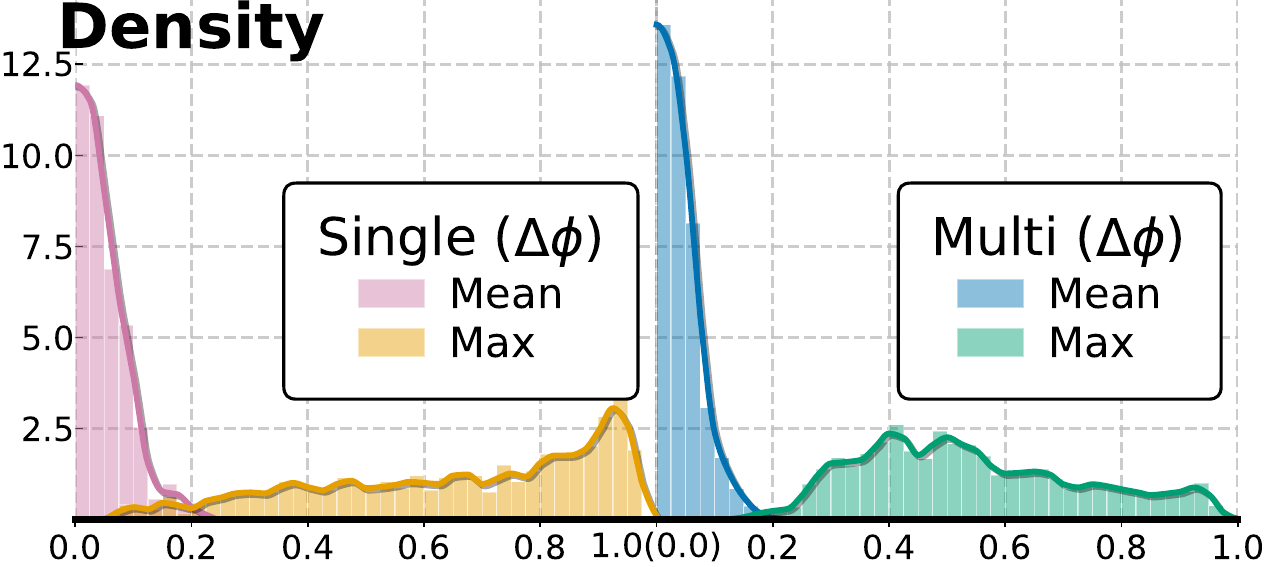}
\vspace{-15pt}
\caption{\textbf{The \(\Delta\phi\) probability density function} for both single- and multi-object setups, reported with groupwise mean and max measurements. Higher density indicates a larger proportion of training samples attaining the corresponding \(\Delta\phi\) level.}
    \label{fig:pdf}
\vspace{-12pt}
\end{figure}

%% file: tables/ablation_studies/hallucination.tex
\begin{table}[t]
\caption{\textbf{Hallucination cases} on the ReasonSeg~\cite{lai2024lisa} val. and test splits (with 995 cases in total). The evaluation is conducted on CoT thinking process using an \texttt{LLM-as-a-Judge} protocol with GPT-5. More details are provided in the  Supp. Section \textcolor{red}{C}.}
\centering
\renewcommand{\arraystretch}{1.1}
\vspace{-10pt}
\label{tab:failure_types}
\resizebox{.87\linewidth}{!}{
\begin{tabular}{?r?ccc?}
\specialrule{1.2pt}{0pt}{0pt}
\multicolumn{1}{?c?}{\multirow{2}{*}{Method}} & \multicolumn{3}{c?}{Hallucination cases (\#)} \\
\cline{2-4}
                        & Off Topic & Mismatch & Vague Claim\\
\specialrule{1.2pt}{0pt}{0pt}
SegZero-7B~\cite{liu2025seg}          &      37     &    231      &   52     \\
VisionReasoner~\cite{liu2025visionreasoner}          &    46     &    194    &  82   \\
Ours (single)                   &    22     &    187      &  35     \\
Ours (multi)                   &  24        &      142    &  48     \\
\specialrule{1.2pt}{0pt}{0pt}
\end{tabular}}
\vspace{-10pt}
\end{table}

%% file: sec/5_conclusion.tex
\section{Conclusion}
In this paper, we introduce a group revision  paradigm designed to recover learning signals from hard cases that are overlooked by GRPO. Our approach begins by sampling an initial response and generating a group of revised results conditioned on it. We then introduce a consolidation process that quantifies the improvement of each revised output, yielding a shaping signal used for both reward calculation and advantage scaling. This signal captures the specific improvements introduced by the revision, providing the policy with concrete guidance on which changes led to the corrected outcome.
Experiments show that our method consistently improves object-level grounding over previous approaches. 
These results highlight its effectiveness as a more reliable optimisation paradigm for LVLMs grounding.

%% file: main.bib
@String(CVPR= {IEEE Conf. Comput. Vis. Pattern Recog.})

@String(ECCV= {Eur. Conf. Comput. Vis.})

@String(CVPR  = {CVPR})

@String(ECCV  = {ECCV})

@article{alayrac2022flamingo,
  title={Flamingo: a visual language model for few-shot learning},
  author={Alayrac, Jean-Baptiste and Donahue, Jeff and Luc, Pauline and Miech, Antoine and Barr, Iain and Hasson, Yana and Lenc, Karel and Mensch, Arthur and Millican, Katherine and Reynolds, Malcolm and others},
  journal={Advances in neural information processing systems},
  volume={35},
  pages={23716--23736},
  year={2022}
}

@inproceedings{li2022blip,
  title={Blip: Bootstrapping language-image pre-training for unified vision-language understanding and generation},
  author={Li, Junnan and Li, Dongxu and Xiong, Caiming and Hoi, Steven},
  booktitle={International conference on machine learning},
  pages={12888--12900},
  year={2022},
  organization={PMLR}
}

@inproceedings{li2023blip,
  title={Blip-2: Bootstrapping language-image pre-training with frozen image encoders and large language models},
  author={Li, Junnan and Li, Dongxu and Savarese, Silvio and Hoi, Steven},
  booktitle={International conference on machine learning},
  pages={19730--19742},
  year={2023},
  organization={PMLR}
}

@misc{wu2025qwenimagetechnicalreport,
      title={Qwen-Image Technical Report}, 
      author={Chenfei Wu and Jiahao Li and Jingren Zhou and Junyang Lin and Kaiyuan Gao and Kun Yan and Sheng-ming Yin and Shuai Bai and Xiao Xu and Yilei Chen and Yuxiang Chen and Zecheng Tang and Zekai Zhang and Zhengyi Wang and An Yang and Bowen Yu and Chen Cheng and Dayiheng Liu and Deqing Li and Hang Zhang and Hao Meng and Hu Wei and Jingyuan Ni and Kai Chen and Kuan Cao and Liang Peng and Lin Qu and Minggang Wu and Peng Wang and Shuting Yu and Tingkun Wen and Wensen Feng and Xiaoxiao Xu and Yi Wang and Yichang Zhang and Yongqiang Zhu and Yujia Wu and Yuxuan Cai and Zenan Liu},
      year={2025},
      eprint={2508.02324},
      archivePrefix={arXiv},
      primaryClass={cs.CV},
      url={https://arxiv.org/abs/2508.02324}, 
}

@article{hsieh2023sugarcrepe,
  title={Sugarcrepe: Fixing hackable benchmarks for vision-language compositionality},
  author={Hsieh, Cheng-Yu and Zhang, Jieyu and Ma, Zixian and Kembhavi, Aniruddha and Krishna, Ranjay},
  journal={Advances in neural information processing systems},
  volume={36},
  pages={31096--31116},
  year={2023}
}

@article{liu2023visual,
  title={Visual instruction tuning},
  author={Liu, Haotian and Li, Chunyuan and Wu, Qingyang and Lee, Yong Jae},
  journal={Advances in neural information processing systems},
  volume={36},
  pages={34892--34916},
  year={2023}
}

@article{zhu2023minigpt,
  title={Minigpt-4: Enhancing vision-language understanding with advanced large language models},
  author={Zhu, Deyao and Chen, Jun and Shen, Xiaoqian and Li, Xiang and Elhoseiny, Mohamed},
  journal={arXiv preprint arXiv:2304.10592},
  year={2023}
}

@article{wang2025mixture,
  title={Mixture of Gaussian-Distributed Prototypes with Generative Modelling for Interpretable and Trustworthy Image Recognition},
  author={Wang, Chong and Chen, Yuanhong and Liu, Fengbei and Liu, Yuyuan and McCarthy, Davis James and Frazer, Helen and Carneiro, Gustavo},
  journal={IEEE Transactions on Pattern Analysis and Machine Intelligence},
  year={2025},
  publisher={IEEE}
}

@article{driess2023palm,
  title={Palm-e: An embodied multimodal language model},
  author={Driess, Danny and Xia, Fei and Sajjadi, Mehdi SM and Lynch, Corey and Chowdhery, Aakanksha and Ichter, Brian and Wahid, Ayzaan and Tompson, Jonathan and Vuong, Quan and Yu, Tianhe and others},
  journal={arXiv preprint arXiv:2303.03378},
  year={2023}
}

@article{chen2025janus,
  title={Janus-Pro: Unified Multimodal Understanding and Generation with Data and Model Scaling},
  author={Chen, Xiaokang and Wu, Zhiyu and Liu, Xingchao and Pan, Zizheng and Liu, Wen and Xie, Zhenda and Yu, Xingkai and Ruan, Chong},
  journal={arXiv preprint arXiv:2501.17811},
  year={2025}
}

@article{comanici2025gemini,
  title={Gemini 2.5: Pushing the frontier with advanced reasoning, multimodality, long context, and next generation agentic capabilities},
  author={Comanici, Gheorghe and Bieber, Eric and Schaekermann, Mike and Pasupat, Ice and Sachdeva, Noveen and Dhillon, Inderjit and Blistein, Marcel and Ram, Ori and Zhang, Dan and Rosen, Evan and others},
  journal={arXiv preprint arXiv:2507.06261},
  year={2025}
}

@article{bai2025qwen2,
  title={Qwen2. 5-vl technical report},
  author={Bai, Shuai and Chen, Keqin and Liu, Xuejing and Wang, Jialin and Ge, Wenbin and Song, Sibo and Dang, Kai and Wang, Peng and Wang, Shijie and Tang, Jun and others},
  journal={arXiv preprint arXiv:2502.13923},
  year={2025}
}

@article{wang2024qwen2,
  title={Qwen2-vl: Enhancing vision-language model's perception of the world at any resolution},
  author={Wang, Peng and Bai, Shuai and Tan, Sinan and Wang, Shijie and Fan, Zhihao and Bai, Jinze and Chen, Keqin and Liu, Xuejing and Wang, Jialin and Ge, Wenbin and others},
  journal={arXiv preprint arXiv:2409.12191},
  year={2024}
}

@article{chen2024expanding,
  title={Expanding performance boundaries of open-source multimodal models with model, data, and test-time scaling},
  author={Chen, Zhe and Wang, Weiyun and Cao, Yue and Liu, Yangzhou and Gao, Zhangwei and Cui, Erfei and Zhu, Jinguo and Ye, Shenglong and Tian, Hao and Liu, Zhaoyang and others},
  journal={arXiv preprint arXiv:2412.05271},
  year={2024}
}

@article{zhu2025internvl3,
  title={Internvl3: Exploring advanced training and test-time recipes for open-source multimodal models},
  author={Zhu, Jinguo and Wang, Weiyun and Chen, Zhe and Liu, Zhaoyang and Ye, Shenglong and Gu, Lixin and Tian, Hao and Duan, Yuchen and Su, Weijie and Shao, Jie and others},
  journal={arXiv preprint arXiv:2504.10479},
  year={2025}
}

@article{wang2025visualprm,
  title={Visualprm: An effective process reward model for multimodal reasoning},
  author={Wang, Weiyun and Gao, Zhangwei and Chen, Lianjie and Chen, Zhe and Zhu, Jinguo and Zhao, Xiangyu and Liu, Yangzhou and Cao, Yue and Ye, Shenglong and Zhu, Xizhou and others},
  journal={arXiv preprint arXiv:2503.10291},
  year={2025}
}

@inproceedings{he2024ma,
  title={Ma-lmm: Memory-augmented large multimodal model for long-term video understanding},
  author={He, Bo and Li, Hengduo and Jang, Young Kyun and Jia, Menglin and Cao, Xuefei and Shah, Ashish and Shrivastava, Abhinav and Lim, Ser-Nam},
  booktitle={Proceedings of the IEEE/CVF Conference on Computer Vision and Pattern Recognition},
  pages={13504--13514},
  year={2024}
}

@inproceedings{lai2024lisa,
  title={Lisa: Reasoning segmentation via large language model},
  author={Lai, Xin and Tian, Zhuotao and Chen, Yukang and Li, Yanwei and Yuan, Yuhui and Liu, Shu and Jia, Jiaya},
  booktitle={Proceedings of the IEEE/CVF Conference on Computer Vision and Pattern Recognition},
  pages={9579--9589},
  year={2024}
}

@inproceedings{xia2024gsva,
  title={Gsva: Generalized segmentation via multimodal large language models},
  author={Xia, Zhuofan and Han, Dongchen and Han, Yizeng and Pan, Xuran and Song, Shiji and Huang, Gao},
  booktitle={Proceedings of the IEEE/CVF Conference on Computer Vision and Pattern Recognition},
  pages={3858--3869},
  year={2024}
}

@inproceedings{rasheed2024glamm,
  title={Glamm: Pixel grounding large multimodal model},
  author={Rasheed, Hanoona and Maaz, Muhammad and Shaji, Sahal and Shaker, Abdelrahman and Khan, Salman and Cholakkal, Hisham and Anwer, Rao M and Xing, Eric and Yang, Ming-Hsuan and Khan, Fahad S},
  booktitle={Proceedings of the IEEE/CVF Conference on Computer Vision and Pattern Recognition},
  pages={13009--13018},
  year={2024}
}

@article{wu2024visionllm,
  title={Visionllm v2: An end-to-end generalist multimodal large language model for hundreds of vision-language tasks},
  author={Wu, Jiannan and Zhong, Muyan and Xing, Sen and Lai, Zeqiang and Liu, Zhaoyang and Chen, Zhe and Wang, Wenhai and Zhu, Xizhou and Lu, Lewei and Lu, Tong and others},
  journal={Advances in Neural Information Processing Systems},
  volume={37},
  pages={69925--69975},
  year={2024}
}

@inproceedings{bai2024videolisa,
  title={One Token to Seg Them All: Language-Instructed Reasoning Segmentation in Videos},
  author={Bai, Zechen and He, Tong and Mei, Haiyang and others},
  booktitle={NeurIPS},
  year={2024},
}

@inproceedings{yan2024visa,
  title={VISA: Reasoning Video Object Segmentation via Large Language Models},
  author={Yan, Cilin and Wang, Haochen and Yan, Shilin and others},
  booktitle={ECCV},
  year={2024},
}

@article{yu2025vpt,
  title={Introducing Visual Perception Token into Multimodal Large Language Model},
  author={Yu, Runpeng and Ma, Xinyin and Wang, Xinchao},
  journal={arXiv:2502.17425},
  year={2025},
  url={https://arxiv.org/abs/2502.17425}
}

@inproceedings{bigverdi2025perceptiontokens,
  title={Perception Tokens Enhance Visual Reasoning in Multimodal Language Models},
  author={Bigverdi, Mohammad and Singh, Amanpreet and others},
  booktitle={CVPR},
  year={2025},
}

@article{li2024llava,
  title={Llava-onevision: Easy visual task transfer},
  author={Li, Bo and Zhang, Yuanhan and Guo, Dong and Zhang, Renrui and Li, Feng and Zhang, Hao and Zhang, Kaichen and Zhang, Peiyuan and Li, Yanwei and Liu, Ziwei and others},
  journal={arXiv preprint arXiv:2408.03326},
  year={2024}
}

@inproceedings{qian2024reasoning,
  title={Reasoning to Attend: Try to Understand How< SEG> Token Works},
  author={Qian, Rui and Yin, Xin and Dou, Dejing},
  booktitle={Proceedings of the IEEE/CVF Conference on Computer Vision and Pattern Recognition},
  year={2025}
}

@article{wang2024segllm,
  title={Segllm: Multi-round reasoning segmentation},
  author={Wang, XuDong and Zhang, Shaolun and Li, Shufan and Kallidromitis, Konstantinos and Li, Kehan and Kato, Yusuke and Kozuka, Kazuki and Darrell, Trevor},
  journal={arXiv preprint arXiv:2410.18923},
  year={2024}
}

@inproceedings{ren2024pixellm,
  title={Pixellm: Pixel reasoning with large multimodal model},
  author={Ren, Zhongwei and Huang, Zhicheng and Wei, Yunchao and Zhao, Yao and Fu, Dongmei and Feng, Jiashi and Jin, Xiaojie},
  booktitle={Proceedings of the IEEE/CVF Conference on Computer Vision and Pattern Recognition},
  pages={26374--26383},
  year={2024}
}

@article{zhou2025reinforcedMLLM,
  title        = {Reinforced MLLM: A Survey on RL‑Based Reasoning in Multimodal Large Language Models},
  author       = {Zhou, Guanghao and Qiu, Panjia and Chen, Cen and Wang, Jie and Yang, Zheming and Xu, Jian and Qiu, Minghui},
  journal      = {arXiv preprint arXiv:2504.21277},
  year         = {2025},
  url          = {https://arxiv.org/abs/2504.21277}
}

@inproceedings{ouyang2022instructgpt,
  title     = {Training Language Models to Follow Instructions with Human Feedback},
  author    = {Ouyang, Long and Wu, Jeff and Jiang, Xu and Almeida, Diogo and Wainwright, Carroll and Mishkin, Pamela and Zhang, Chong and Agarwal, Sandhini and Slama, Katarina and Ray, Alex and others},
  booktitle = {Advances in Neural Information Processing Systems (NeurIPS)},
  year      = {2022},
  url       = {https://proceedings.neurips.cc/paper_files/paper/2022/file/b1efde53be364a73914f58805a001731-Paper-Conference.pdf},
  eprint    = {2203.02155},
  archivePrefix = {arXiv}
}

@article{shao2024grpo,
  title   = {DeepSeekMath: Pushing the Limits of Mathematical Reasoning in Open Language Models},
  author  = {Shao, Zhihong and Zhang, Zihan and Huang, Xinyu and Yang, Yushi and Qiu, Minghui and Zhao, Wayne Xin},
  journal = {arXiv preprint arXiv:2402.03300},
  year    = {2024},
  url     = {https://arxiv.org/abs/2402.03300},
  note    = {Introduces Group-Relative Policy Optimization (GRPO)}
}

@article{guo2025deepseekr1,
  title   = {DeepSeek-R1: Incentivizing Reasoning Capability in LLMs via Reinforcement Learning},
  author  = {Guo, Dong and Liu, Zichen and Zhang, Weize and Zhou, Yuxuan and Wang, Xiaozhi and others},
  journal = {arXiv preprint arXiv:2501.12948},
  year    = {2025},
  url     = {https://arxiv.org/abs/2501.12948},
  note    = {Uses GRPO for reasoning RL}
}

@article{yu2025dapo,
  title   = {DAPO: Decoupled Clip and Dynamic Sampling Policy Optimization for Large Language Model Reinforcement Learning},
  author  = {Yu, Qiyuan and Liu, Jue and Li, Xiaoyu and Xu, Yicheng and Zhang, Tianyi and others},
  journal = {arXiv preprint arXiv:2503.14476},
  year    = {2025},
  url     = {https://arxiv.org/abs/2503.14476}
}

@article{bai2025univg,
  title={UniVG-R1: Reasoning guided universal visual grounding with reinforcement learning},
  author={Bai, Sule and Li, Mingxing and Liu, Yong and Tang, Jing and Zhang, Haoji and Sun, Lei and Chu, Xiangxiang and Tang, Yansong},
  journal={arXiv preprint arXiv:2505.14231},
  year={2025},
  url={https://arxiv.org/abs/2505.14231}
}

@article{jiang2025rexthinker,
  title={Rex-Thinker: Grounded object referring via chain-of-thought reasoning},
  author={Jiang, Qing and Chen, Xingyu and Zeng, Zhaoyang and Yu, Junzhi and Zhang, Lei},
  journal={arXiv preprint arXiv:2506.04034},
  year={2025},
  url={https://arxiv.org/abs/2506.04034}
}

@article{xu2025medgroundr1,
  title={MedGround-R1: Advancing medical image grounding via spatial-semantic rewarded group relative policy optimization},
  author={Xu, Huihui and Nie, Yuanpeng and Wang, Hualiang and Chen, Ying and Li, Wei and Ning, Junzhi and Liu, Lihao and Wang, Hongqiu and Zhu, Lei and Liu, Jiyao and Li, Xiaomeng and He, Junjun},
  journal={arXiv preprint arXiv:2507.02994},
  year={2025},
  url={https://arxiv.org/abs/2507.02994}
}

@article{zhou2025guig1,
  title={GUI-G1: Understanding R1-Zero-like training for visual grounding in GUI agents},
  author={Zhou, Yuqi and Dai, Sunhao and Wang, Shuai and Zhou, Kaiwen and Jia, Qinglin and Xu, Jun},
  journal={arXiv preprint arXiv:2505.15810},
  year={2025},
  url={https://arxiv.org/abs/2505.15810}
}

@article{ye2025guiarp,
  title={GUI-ARP: Enhancing grounding with adaptive region perception for GUI agents},
  author={Ye, Xianhang and Li, Yiqing and Dai, Wei and Liu, Miancan and Chen, Ziyuan and Han, Zhangye and Min, Hongbo and Ren, Jinkui and Zhang, Xiantao and Yang, Wen and Jin, Zhi},
  journal={arXiv preprint arXiv:2509.15532},
  year={2025},
  url={https://arxiv.org/abs/2509.15532}
}

@article{yang2025look,
  title={Look-back: Implicit visual re-focusing in mllm reasoning},
  author={Yang, Shuo and Niu, Yuwei and Liu, Yuyang and Ye, Yang and Lin, Bin and Yuan, Li},
  journal={arXiv preprint arXiv:2507.03019},
  year={2025}
}

@article{cao2025groundr1,
  title={Ground-R1: Incentivizing grounded visual reasoning via reinforcement learning},
  author={Cao, Meng and Zhao, Haoze and Zhang, Can and Chang, Xiaojun and Reid, Ian and Liang, Xiaodan},
  journal={arXiv preprint arXiv:2505.20272},
  year={2025},
  url={https://arxiv.org/abs/2505.20272}
}

@article{pan2025medvlmr1,
  title={MedVLM-R1: Incentivizing medical reasoning capability of vision-language models via reinforcement learning},
  author={Pan, Jiazhen and Liu, Che and Wu, Junde and Liu, Fenglin and Zhu, Jiayuan and Li, Hongwei Bran and Chen, Chen and Ouyang, Cheng and Rueckert, Daniel},
  journal={arXiv preprint arXiv:2502.19634},
  year={2025},
  url={https://arxiv.org/abs/2502.19634}
}

@article{shen2025vlm,
  title={Vlm-r1: A stable and generalizable r1-style large vision-language model},
  author={Shen, Haozhan and Liu, Peng and Li, Jingcheng and Fang, Chunxin and Ma, Yibo and Liao, Jiajia and Shen, Qiaoli and Zhang, Zilun and Zhao, Kangjia and Zhang, Qianqian and Xu, Ruochen and Zhao, Tiancheng },
  journal={arXiv preprint arXiv:2504.07615},
  year={2025}
}

@article{you2025segr1,
  title={Seg-R1: Segmentation can be surprisingly simple with reinforcement learning},
  author={You, Zuyao and Wu, Zuxuan},
  journal={arXiv preprint arXiv:2506.22624},
  year={2025},
  url={https://arxiv.org/abs/2506.22624}
}

@article{liu2025seg,
  title={Seg-zero: Reasoning-chain guided segmentation via cognitive reinforcement},
  author={Liu, Yuqi and Peng, Bohao and Zhong, Zhisheng and Yue, Zihao and Lu, Fanbin and Yu, Bei and Jia, Jiaya},
  journal={arXiv preprint arXiv:2503.06520},
  year={2025}
}

@article{huang2025samr1,
  title={SAM-R1: Leveraging SAM for reward feedback in multimodal segmentation via reinforcement learning},
  author={Huang, Jiaqi and Xu, Zunnan and Zhou, Jun and Liu, Ting and Xiao, Yicheng and Ou, Mingwen and Ji, Bowen and Li, Xiu and Yuan, Kehong},
  journal={arXiv preprint arXiv:2505.22596},
  year={2025},
  url={https://arxiv.org/abs/2505.22596}
}

@article{liu2025visionreasoner,
  title={VisionReasoner: Unified visual perception and reasoning via reinforcement learning},
  author={Liu, Yuqi and Qu, Tianyuan and Zhong, Zhisheng and Peng, Bohao and Liu, Shu and Yu, Bei and Jia, Jiaya},
  journal={arXiv preprint arXiv:2505.12081},
  year={2025},
  url={https://arxiv.org/abs/2505.12081}
}

@inproceedings{liu2023gres,
  title={Gres: Generalized referring expression segmentation},
  author={Liu, Chang and Ding, Henghui and Jiang, Xudong},
  booktitle={Proceedings of the IEEE/CVF conference on computer vision and pattern recognition},
  pages={23592--23601},
  year={2023}
}

@inproceedings{kirillov2023segment,
  title={Segment anything},
  author={Kirillov, Alexander and Mintun, Eric and Ravi, Nikhila and Mao, Hanzi and Rolland, Chloe and Gustafson, Laura and Xiao, Tete and Whitehead, Spencer and Berg, Alexander C and Lo, Wan-Yen and others},
  booktitle={Proceedings of the IEEE/CVF international conference on computer vision},
  pages={4015--4026},
  year={2023}
}

@article{ravi2024sam2,
  title={SAM 2: Segment Anything in Images and Videos},
  author={Ravi, Nikhila and Gabeur, Valentin and Hu, Yuan-Ting and Hu, Ronghang and Ryali, Chaitanya and Ma, Tengyu and Khedr, Haitham and R{\"a}dle, Roman and Rolland, Chloe and Gustafson, Laura and Mintun, Eric and Pan, Junting and Alwala, Kalyan Vasudev and Carion, Nicolas and Wu, Chao-Yuan and Girshick, Ross and Doll{\'a}r, Piotr and Feichtenhofer, Christoph},
  journal={arXiv preprint arXiv:2408.00714},
  year={2024},
  url={https://arxiv.org/abs/2408.00714}
}

@inproceedings{ng1999policy,
  title={Policy invariance under reward transformations: Theory and application to reward shaping},
  author={Ng, Andrew Y and Harada, Daishi and Russell, Stuart},
  booktitle={Icml},
  volume={99},
  pages={278--287},
  year={1999},
  organization={Citeseer}
}

@article{liu2024survey,
  title={A survey on hallucination in large vision-language models},
  author={Liu, Hanchao and Xue, Wenyuan and Chen, Yifei and Chen, Dapeng and Zhao, Xiutian and Wang, Ke and Hou, Liping and Li, Rongjun and Peng, Wei},
  journal={arXiv preprint arXiv:2402.00253},
  year={2024}
}

@inproceedings{leng2024mitigating,
  title={Mitigating object hallucinations in large vision-language models through visual contrastive decoding},
  author={Leng, Sicong and Zhang, Hang and Chen, Guanzheng and Li, Xin and Lu, Shijian and Miao, Chunyan and Bing, Lidong},
  booktitle={Proceedings of the IEEE/CVF Conference on Computer Vision and Pattern Recognition},
  pages={13872--13882},
  year={2024}
}

@article{rohrbach2018object,
  title={Object hallucination in image captioning},
  author={Rohrbach, Anna and Hendricks, Lisa Anne and Burns, Kaylee and Darrell, Trevor and Saenko, Kate},
  journal={arXiv preprint arXiv:1809.02156},
  year={2018}
}

@inproceedings{biten2022let,
  title={Let there be a clock on the beach: Reducing object hallucination in image captioning},
  author={Biten, Ali Furkan and G{\'o}mez, Llu{\'\i}s and Karatzas, Dimosthenis},
  booktitle={Proceedings of the IEEE/CVF Winter Conference on Applications of Computer Vision},
  pages={1381--1390},
  year={2022}
}

@article{ji2023survey,
  title={Survey of hallucination in natural language generation},
  author={Ji, Ziwei and Lee, Nayeon and Frieske, Rita and Yu, Tiezheng and Su, Dan and Xu, Yan and Ishii, Etsuko and Bang, Ye Jin and Madotto, Andrea and Fung, Pascale},
  journal={ACM computing surveys},
  volume={55},
  number={12},
  pages={1--38},
  year={2023},
  publisher={ACM New York, NY}
}

@article{bai2024hallucination,
  title={Hallucination of multimodal large language models: A survey},
  author={Bai, Zechen and Wang, Pichao and Xiao, Tianjun and He, Tong and Han, Zongbo and Zhang, Zheng and Shou, Mike Zheng},
  journal={arXiv preprint arXiv:2404.18930},
  year={2024}
}

@inproceedings{gupta2019lvis,
  title={Lvis: A dataset for large vocabulary instance segmentation},
  author={Gupta, Agrim and Dollar, Piotr and Girshick, Ross},
  booktitle={Proceedings of the IEEE/CVF conference on computer vision and pattern recognition},
  pages={5356--5364},
  year={2019}
}

@inproceedings{pi2024perceptiongpt,
  title={Perceptiongpt: Effectively fusing visual perception into llm},
  author={Pi, Renjie and Yao, Lewei and Gao, Jiahui and Zhang, Jipeng and Zhang, Tong},
  booktitle={Proceedings of the IEEE/CVF conference on computer vision and pattern recognition},
  pages={27124--27133},
  year={2024}
}

@article{liu2025visual,
  title={Visual-rft: Visual reinforcement fine-tuning},
  author={Liu, Ziyu and Sun, Zeyi and Zang, Yuhang and Dong, Xiaoyi and Cao, Yuhang and Duan, Haodong and Lin, Dahua and Wang, Jiaqi},
  journal={arXiv preprint arXiv:2503.01785},
  year={2025}
}

@article{zang2024overcoming,
  title={Overcoming the pitfalls of vision-language model finetuning for ood generalization},
  author={Zang, Yuhang and Goh, Hanlin and Susskind, Josh and Huang, Chen},
  journal={arXiv preprint arXiv:2401.15914},
  year={2024}
}

@article{zhang2024vision,
  title={Vision-language models for vision tasks: A survey},
  author={Zhang, Jingyi and Huang, Jiaxing and Jin, Sheng and Lu, Shijian},
  journal={IEEE transactions on pattern analysis and machine intelligence},
  volume={46},
  number={8},
  pages={5625--5644},
  year={2024},
  publisher={IEEE}
}

@article{chung2024scaling,
  title={Scaling instruction-finetuned language models},
  author={Chung, Hyung Won and Hou, Le and Longpre, Shayne and Zoph, Barret and Tay, Yi and Fedus, William and Li, Yunxuan and Wang, Xuezhi and Dehghani, Mostafa and Brahma, Siddhartha and others},
  journal={Journal of Machine Learning Research},
  volume={25},
  number={70},
  pages={1--53},
  year={2024}
}

@inproceedings{yu2016modeling,
  title={Modeling context in referring expressions},
  author={Yu, Licheng and Poirson, Patrick and Yang, Shan and Berg, Alexander C and Berg, Tamara L},
  booktitle={European conference on computer vision},
  pages={69--85},
  year={2016},
  organization={Springer}
}

@article{yang2023lisa++,
  title={Lisa++: An improved baseline for reasoning segmentation with large language model},
  author={Yang, Senqiao and Qu, Tianyuan and Lai, Xin and Tian, Zhuotao and Peng, Bohao and Liu, Shu and Jia, Jiaya},
  journal={arXiv preprint arXiv:2312.17240},
  year={2023}
}

@inproceedings{paiss2023teaching,
  title={Teaching clip to count to ten},
  author={Paiss, Roni and Ephrat, Ariel and Tov, Omer and Zada, Shiran and Mosseri, Inbar and Irani, Michal and Dekel, Tali},
  booktitle={Proceedings of the IEEE/CVF International Conference on Computer Vision},
  pages={3170--3180},
  year={2023}
}

@inproceedings{deitke2025molmo,
  title={Molmo and pixmo: Open weights and open data for state-of-the-art vision-language models},
  author={Deitke, Matt and Clark, Christopher and Lee, Sangho and Tripathi, Rohun and Yang, Yue and Park, Jae Sung and Salehi, Mohammadreza and Muennighoff, Niklas and Lo, Kyle and Soldaini, Luca and others},
  booktitle={Proceedings of the Computer Vision and Pattern Recognition Conference},
  pages={91--104},
  year={2025}
}

@article{lightman2023lets,
      title={Let's Verify Step by Step}, 
      author={Lightman, Hunter and Kosaraju, Vineet and Burda, Yura and Edwards, Harri and Baker, Bowen and Lee, Teddy and Leike, Jan and Schulman, John and Sutskever, Ilya and Cobbe, Karl},
      journal={arXiv preprint arXiv:2305.20050},
      year={2023}
}

@inproceedings{fu2025mme,
  title={MME: A comprehensive evaluation benchmark for multimodal large language models},
  author={Fu, Chaoyou and Chen, Peixian and Shen, Yunhang and Qin, Yulei and Zhang, Mengdan and Lin, Xu and Yang, Jinrui and Zheng, Xiawu and Li, Ke and Sun, Xing and others},
  booktitle={The Thirty-ninth Annual Conference on Neural Information Processing Systems Datasets and Benchmarks Track},
  year={2025}
}

@inproceedings{yue2025mmmu,
  title={Mmmu-pro: A more robust multi-discipline multimodal understanding benchmark},
  author={Yue, Xiang and Zheng, Tianyu and Ni, Yuansheng and Wang, Yubo and Zhang, Kai and Tong, Shengbang and Sun, Yuxuan and Yu, Botao and Zhang, Ge and Sun, Huan and others},
  booktitle={Proceedings of the 63rd Annual Meeting of the Association for Computational Linguistics (Volume 1: Long Papers)},
  pages={15134--15186},
  year={2025}
}

@article{li2023evaluating,
  title={Evaluating object hallucination in large vision-language models},
  author={Li, Yifan and Du, Yifan and Zhou, Kun and Wang, Jinpeng and Zhao, Wayne Xin and Wen, Ji-Rong},
  journal={arXiv preprint arXiv:2305.10355},
  year={2023}
}

@article{liu2024ocrbench,
  title={Ocrbench: on the hidden mystery of ocr in large multimodal models},
  author={Liu, Yuliang and Li, Zhang and Huang, Mingxin and Yang, Biao and Yu, Wenwen and Li, Chunyuan and Yin, Xu-Cheng and Liu, Cheng-Lin and Jin, Lianwen and Bai, Xiang},
  journal={Science China Information Sciences},
  volume={67},
  number={12},
  pages={220102},
  year={2024},
  publisher={Springer}
}

@inproceedings{masry2022chartqa,
  title={Chartqa: A benchmark for question answering about charts with visual and logical reasoning},
  author={Masry, Ahmed and Do, Xuan Long and Tan, Jia Qing and Joty, Shafiq and Hoque, Enamul},
  booktitle={Findings of the association for computational linguistics: ACL 2022},
  pages={2263--2279},
  year={2022}
}

@article{li2024seed,
  title={Seed-bench-2-plus: Benchmarking multimodal large language models with text-rich visual comprehension},
  author={Li, Bohao and Ge, Yuying and Chen, Yi and Ge, Yixiao and Zhang, Ruimao and Shan, Ying},
  journal={arXiv preprint arXiv:2404.16790},
  year={2024}
}

@article{zhang2025lessons,
  title={The lessons of developing process reward models in mathematical reasoning},
  author={Zhang, Zhenru and Zheng, Chujie and Wu, Yangzhen and Zhang, Beichen and Lin, Runji and Yu, Bowen and Liu, Dayiheng and Zhou, Jingren and Lin, Junyang},
  journal={arXiv preprint arXiv:2501.07301},
  year={2025}
}

@article{mroueh2025revisiting,
  title={Revisiting Group Relative Policy Optimization: Insights into On-Policy and Off-Policy Training},
  author={Mroueh, Youssef and Dupuis, Nicolas and Belgodere, Brian and Nitsure, Apoorva and Rigotti, Mattia and Greenewald, Kristjan and Navratil, Jiri and Ross, Jerret and Rios, Jesus},
  journal={arXiv preprint arXiv:2505.22257},
  year={2025}
}

@article{wiewiora2003potential,
  title={Potential-based shaping and Q-value initialization are equivalent},
  author={Wiewiora, Eric},
  journal={Journal of Artificial Intelligence Research},
  volume={19},
  pages={205--208},
  year={2003}
}

@inproceedings{devlin2014potential,
  title={Potential-based difference rewards for multiagent reinforcement learning},
  author={Devlin, Sam and Yliniemi, Logan and Kudenko, Daniel and Tumer, Kagan},
  booktitle={Proceedings of the 2014 international conference on Autonomous agents and multi-agent systems},
  pages={165--172},
  year={2014}
}

@article{christiano2017deep,
  title={Deep reinforcement learning from human preferences},
  author={Christiano, Paul F and Leike, Jan and Brown, Tom and Martic, Miljan and Legg, Shane and Amodei, Dario},
  journal={Advances in neural information processing systems},
  volume={30},
  year={2017}
}

@article{andrychowicz2017hindsight,
  title={Hindsight experience replay},
  author={Andrychowicz, Marcin and Wolski, Filip and Ray, Alex and Schneider, Jonas and Fong, Rachel and Welinder, Peter and McGrew, Bob and Tobin, Josh and Pieter Abbeel, OpenAI and Zaremba, Wojciech},
  journal={Advances in neural information processing systems},
  volume={30},
  year={2017}
}

@inproceedings{chen2024boosting,
  title={Boosting Reinforcement Learning Algorithms in Continuous Robotic Reaching Tasks Using Adaptive Potential Functions},
  author={Chen, Yifei and Schomaker, Lambert and Cruz, Francisco},
  booktitle={Australasian Joint Conference on Artificial Intelligence},
  pages={52--64},
  year={2024},
  organization={Springer}
}

@inproceedings{camacho2021reward,
  title={Reward machines for vision-based robotic manipulation},
  author={Camacho, Alberto and Varley, Jacob and Zeng, Andy and Jain, Deepali and Iscen, Atil and Kalashnikov, Dmitry},
  booktitle={2021 IEEE International Conference on Robotics and Automation (ICRA)},
  pages={14284--14290},
  year={2021},
  organization={IEEE}
}

@inproceedings{devlin2012dynamic,
  title={Dynamic potential-based reward shaping},
  author={Devlin, Sam Michael and Kudenko, Daniel},
  booktitle={11th International Conference on Autonomous Agents and Multiagent Systems (AAMAS 2012)},
  pages={433--440},
  year={2012},
  organization={IFAAMAS}
}

@article{sheng2024hybridflow,
  title   = {HybridFlow: A Flexible and Efficient RLHF Framework},
  author  = {Guangming Sheng and Chi Zhang and Zilingfeng Ye and Xibin Wu and Wang Zhang and Ru Zhang and Yanghua Peng and Haibin Lin and Chuan Wu},
  year    = {2024},
  journal = {arXiv preprint arXiv: 2409.19256}
}

@inproceedings{kwon2023efficient,
  title={Efficient Memory Management for Large Language Model Serving with PagedAttention},
  author={Woosuk Kwon and Zhuohan Li and Siyuan Zhuang and Ying Sheng and Lianmin Zheng and Cody Hao Yu and Joseph E. Gonzalez and Hao Zhang and Ion Stoica},
  booktitle={Proceedings of the ACM SIGOPS 29th Symposium on Operating Systems Principles},
  year={2023}
}

@article{yue2024mmmu,
  title={Mmmu-pro: A more robust multi-discipline multimodal understanding benchmark},
  author={Yue, Xiang and Zheng, Tianyu and Ni, Yuansheng and Wang, Yubo and Zhang, Kai and Tong, Shengbang and Sun, Yuxuan and Yu, Botao and Zhang, Ge and Sun, Huan and others},
  journal={arXiv preprint arXiv:2409.02813},
  year={2024}
}

@article{dai2024process,
  title={Process supervision-guided policy optimization for code generation},
  author={Dai, Ning and Wu, Zheng and Zheng, Renjie and Wei, Ziyun and Shi, Wenlei and Jin, Xing and Liu, Guanlin and Dun, Chen and Huang, Liang and Yan, Lin},
  journal={arXiv preprint arXiv:2410.17621},
  year={2024}
}

@article{zheng2025survey,
  title={A Survey of Process Reward Models: From Outcome Signals to Process Supervisions for Large Language Models},
  author={Zheng, Congming and Zhu, Jiachen and Ou, Zhuoying and Chen, Yuxiang and Zhang, Kangning and Shan, Rong and Zheng, Zeyu and Yang, Mengyue and Lin, Jianghao and Yu, Yong and others},
  journal={arXiv preprint arXiv:2510.08049},
  year={2025}
}

@inproceedings{li2025codeprm,
  title={Codeprm: Execution feedback-enhanced process reward model for code generation},
  author={Li, Qingyao and Dai, Xinyi and Li, Xiangyang and Zhang, Weinan and Wang, Yasheng and Tang, Ruiming and Yu, Yong},
  booktitle={Findings of the Association for Computational Linguistics: ACL 2025},
  pages={8169--8182},
  year={2025}
}

@article{she2025r,
  title={R-prm: Reasoning-driven process reward modeling},
  author={She, Shuaijie and Liu, Junxiao and Liu, Yifeng and Chen, Jiajun and Huang, Xin and Huang, Shujian},
  journal={arXiv preprint arXiv:2503.21295},
  year={2025}
}

@article{li2024process,
  title={Process reward model with q-value rankings},
  author={Li, Wendi and Li, Yixuan},
  journal={arXiv preprint arXiv:2410.11287},
  year={2024}
}

@article{yin2025dynamic,
  title={Dynamic and generalizable process reward modeling},
  author={Yin, Zhangyue and Sun, Qiushi and Zeng, Zhiyuan and Cheng, Qinyuan and Qiu, Xipeng and Huang, Xuanjing},
  journal={arXiv preprint arXiv:2507.17849},
  year={2025}
}

@inproceedings{huang2025roboground,
  title={RoboGround: Robotic Manipulation with Grounded Vision-Language Priors},
  author={Huang, Haifeng and Chen, Xinyi and Chen, Yilun and Li, Hao and Han, Xiaoshen and Wang, Zehan and Wang, Tai and Pang, Jiangmiao and Zhao, Zhou},
  booktitle={Proceedings of the Computer Vision and Pattern Recognition Conference},
  pages={22540--22550},
  year={2025}
}

@article{kurz2025benchmarking,
  title={Benchmarking vision-language models for diagnostics in emergency and critical care settings},
  author={Kurz, Christoph F and Merzhevich, Tatiana and Eskofier, Bjoern M and Kather, Jakob Nikolas and Gmeiner, Benjamin},
  journal={npj Digital Medicine},
  volume={8},
  number={1},
  pages={423},
  year={2025},
  publisher={Nature Publishing Group UK London}
}

@inproceedings{wang2025rethinking,
  title={Rethinking the embodied gap in vision-and-language navigation: A holistic study of physical and visual disparities},
  author={Wang, Liuyi and Xia, Xinyuan and Zhao, Hui and Wang, Hanqing and Wang, Tai and Chen, Yilun and Liu, Chengju and Chen, Qijun and Pang, Jiangmiao},
  booktitle={Proceedings of the IEEE/CVF International Conference on Computer Vision},
  pages={9455--9465},
  year={2025}
}

@article{zhang2025chain,
  title={Chain-of-Action: Trajectory Autoregressive Modeling for Robotic Manipulation},
  author={Zhang, Wenbo and Hu, Tianrun and Qiao, Yanyuan and Zhang, Hanbo and Qin, Yuchu and Li, Yang and Liu, Jiajun and Kong, Tao and Liu, Lingqiao and Ma, Xiao},
  journal={arXiv preprint arXiv:2506.09990},
  year={2025}
}

@article{zhao2025manipbench,
  title={ManipBench: Benchmarking vision-language models for low-level robot manipulation},
  author={Zhao, Enyu and Raval, Vedant and Zhang, Hejia and Mao, Jiageng and Shangguan, Zeyu and Nikolaidis, Stefanos and Wang, Yue and Seita, Daniel},
  journal={arXiv preprint arXiv:2505.09698},
  year={2025}
}

@article{qiao2025navbench,
  title={NavBench: Probing Multimodal Large Language Models for Embodied Navigation},
  author={Qiao, Yanyuan and Hong, Haodong and Lyu, Wenqi and An, Dong and Zhang, Siqi and Xie, Yutong and Wang, Xinyu and Wu, Qi},
  journal={arXiv preprint arXiv:2506.01031},
  year={2025}
}

@article{windecker2025navitrace,
  title={NaviTrace: Evaluating Embodied Navigation of Vision-Language Models},
  author={Windecker, Tim and Patel, Manthan and Reuss, Moritz and Schwarzkopf, Richard and Cadena, Cesar and Lioutikov, Rudolf and Hutter, Marco and Frey, Jonas},
  journal={arXiv preprint arXiv:2510.26909},
  year={2025}
}

@inproceedings{pramanick2024jack,
  title={Jack of all tasks master of many: Designing general-purpose coarse-to-fine vision-language model},
  author={Pramanick, Shraman and Han, Guangxing and Hou, Rui and Nag, Sayan and Lim, Ser-Nam and Ballas, Nicolas and Wang, Qifan and Chellappa, Rama and Almahairi, Amjad},
  booktitle={Proceedings of the IEEE/CVF Conference on Computer Vision and Pattern Recognition},
  pages={14076--14088},
  year={2024}
}

@article{feng2025video,
  title={Video-r1: Reinforcing video reasoning in mllms},
  author={Feng, Kaituo and Gong, Kaixiong and Li, Bohao and Guo, Zonghao and Wang, Yibing and Peng, Tianshuo and Wu, Junfei and Zhang, Xiaoying and Wang, Benyou and Yue, Xiangyu},
  journal={arXiv preprint arXiv:2503.21776},
  year={2025}
}

@inproceedings{ma2024groma,
  title={Groma: Localized visual tokenization for grounding multimodal large language models},
  author={Ma, Chuofan and Jiang, Yi and Wu, Jiannan and Yuan, Zehuan and Qi, Xiaojuan},
  booktitle={European Conference on Computer Vision},
  pages={417--435},
  year={2024},
  organization={Springer}
}

@inproceedings{wang2024elysium,
  title={Elysium: Exploring object-level perception in videos via mllm},
  author={Wang, Han and Ye, Yongjie and Wang, Yanjie and Nie, Yuxiang and Huang, Can},
  booktitle={European Conference on Computer Vision},
  pages={166--185},
  year={2024},
  organization={Springer}
}

@misc{zhang2024lmmsevalrealitycheckevaluation,
      title={LMMs-Eval: Reality Check on the Evaluation of Large Multimodal Models}, 
      author={Kaichen Zhang and Bo Li and Peiyuan Zhang and Fanyi Pu and Joshua Adrian Cahyono and Kairui Hu and Shuai Liu and Yuanhan Zhang and Jingkang Yang and Chunyuan Li and Ziwei Liu},
      year={2024},
      eprint={2407.12772},
      archivePrefix={arXiv},
      primaryClass={cs.CL},
      url={https://arxiv.org/abs/2407.12772}, 
}

@article{liu2025auralsam2,
  title={AuralSAM2: Enabling SAM2 Hear Through Pyramid Audio-Visual Feature Prompting},
  author={Liu, Yuyuan and Chen, Yuanhong and Wang, Chong and Han, Junlin and Wu, Junde and Peng, Can and Chen, Jingkun and Tian, Yu and Carneiro, Gustavo},
  journal={arXiv preprint arXiv:2506.01015},
  year={2025}
}
